\documentclass[lettersize,journal]{IEEEtran}
\usepackage{amsmath,amsfonts}
\usepackage{algorithmic}
\usepackage{algorithm}
\usepackage{array}
\usepackage[caption=false,font=normalsize,labelfont=sf,textfont=sf]{subfig}
\usepackage{textcomp}
\usepackage{stfloats}
\usepackage{url}
\usepackage{verbatim}
\usepackage{graphicx}
\usepackage{cite}
\usepackage{color}
\usepackage{bm}
\usepackage{booktabs}
\usepackage{colortbl}
\usepackage{float}
\usepackage{ragged2e}  
\usepackage{caption}
\usepackage{csquotes}
\usepackage[breaklinks=true,bookmarks=false]{hyperref}
\usepackage{graphicx}
\usepackage{amsmath}
\usepackage{amssymb}
\usepackage{booktabs}
\usepackage{multirow}
\usepackage{xcolor}

\begin{document}

\title{CuDi: Curve Distillation for Efficient and Controllable Exposure Adjustment}

\author{Chongyi Li, Chunle Guo, Ruicheng Feng, Shangchen Zhou, and Chen Change Loy,~\IEEEmembership{Senior Member,~IEEE}\\

\thanks{C. Li and C. Guo are the VCIP, CS, Nankai University, Tianjin, China  (e-mail: lichongyi@nankai.edu.cn and guochunle@nankai.edu.cn). R. Feng, S. Zhou, and C. C. Loy are with S-Lab, Nanyang Technological University (NTU), Singapore  (e-mail:  ruicheng002@e.ntu.edu.sg, s200094@e.ntu.edu.sg, and ccloy@ntu.edu.sg).}
\thanks{Partial work was done when C. Li worked at NTU.}
\thanks{C. Li and C. Guo contribute equally.}
\thanks{C. C. Loy is the corresponding author.}
}

\markboth{IEEE TRANSACTIONS ON PATTERN ANALYSIS AND MACHINE INTELLIGENCE}%
{Shell \MakeLowercase{\textit{et al.}}: Bare Demo of IEEEtran.cls for Computer Society Journals}

\maketitle


\begin{abstract}
	We present \textbf{Cu}rve \textbf{Di}stillation, \textbf{CuDi}, for efficient and controllable exposure adjustment without the requirement of paired or unpaired data during training.
	Our method inherits the zero-reference learning and curve-based framework from an effective low-light image enhancement method, Zero-DCE, with further speed up in its inference speed, reduction in its model size, and extension to controllable exposure adjustment.
	The improved inference speed and lightweight model are achieved through novel curve distillation that approximates the time-consuming iterative operation in the conventional curve-based framework by high-order curve's tangent line. 
	The controllable exposure adjustment is made possible with a new self-supervised spatial exposure control loss that constrains the exposure levels of different spatial regions of the output to be close to the brightness distribution of an exposure map serving as an input condition. Different from most existing methods that can only correct either underexposed or overexposed photos, our approach corrects both underexposed and overexposed photos with a single model. Notably, our approach can additionally adjust the exposure levels of a photo globally or locally with the guidance of an input condition exposure map, which can be pre-defined or manually set in the inference stage. Through extensive experiments, we show that our method is appealing for its fast, robust, and flexible performance, outperforming state-of-the-art methods in real scenes. Project page: \url{https://li-chongyi.github.io/CuDi_files/}.
\end{abstract}


\section{Introduction}
\IEEEPARstart{U}nsatisfactory exposure exists in photographs when an inappropriate level of light hits the camera's sensor. 
Typically, underexposed photos are murky and suffer from loss of detail and dull colors while overexposed photos appear washed out and have little detail in their highlights. 
Automatic exposure correction is always desired when one does not have the expertise to adjust the shutter speed, ISO value, and aperture diameter when taking a photo.
It is preferable if the exposure is controllable to allow further adjustment. 

Achieving controllable exposure adjustment is challenging because (1) input photos may have diverse exposure properties such as underexposure, overexposure, and non-uniform exposure; and (2) different exposure adjustments such as global adjustment and local adjustment are needed in different circumstances.
These hurdles challenge the capability of exposure adjustment algorithms.
In comparison to conventional one-time exposure correction, controllable exposure adjustment also demands more flexible operations (e.g., fine-grained adjustment that increases or decreases the exposure levels of a photo globally or locally), faster processing speed (e.g., timely feedback in the interaction process), and more deployable solutions (e.g., lightweight structures).

\begin{figure} [t]
	\begin{center}
		\begin{tabular}{c@{ }}
			\includegraphics[width=0.7\linewidth]{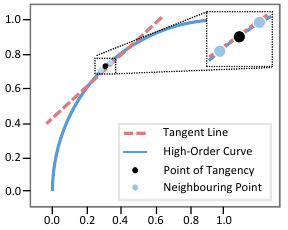}
		\end{tabular}
	\end{center}
	\vspace{-1em}
	\caption{\textbf{Motivation of curve distillation.} A high-order curve can be well approximated by its tangent line near the point of tangency. The tangent line can also inherit the monotonicity relationships between neighboring pixels in the result adjusted by the high-order curve. }
	\vspace{-1.3em}
	\label{fig:motivation}
\end{figure}

\begin{figure*} [t]
	\begin{center}
		\begin{tabular}{c@{ }}
                \includegraphics[width=0.9\linewidth]{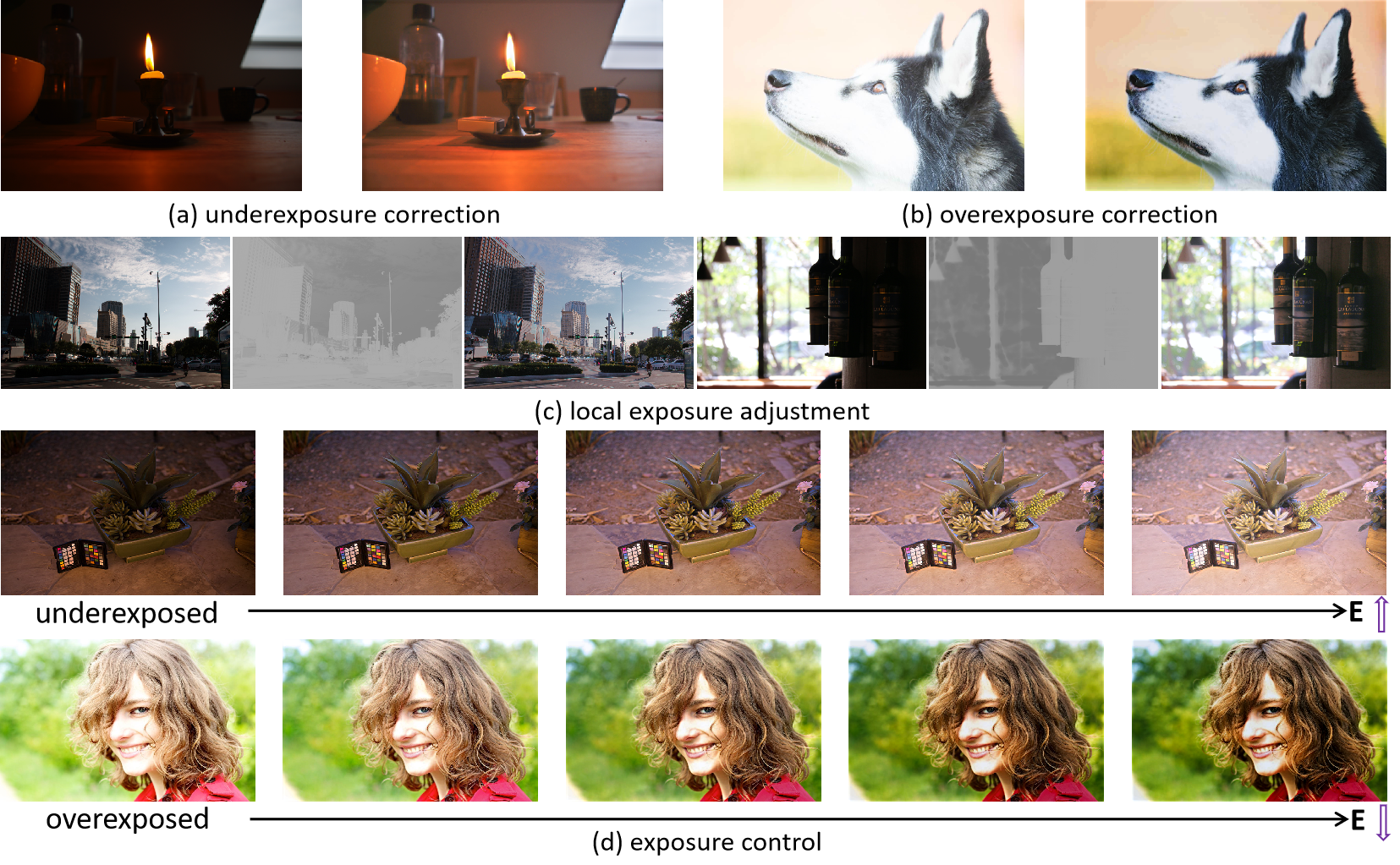}\vspace{0.1cm}
		\end{tabular}
	\end{center}
	\vspace{-2em}
	\caption{\textbf{Results achieved by our method with a single model.} Our method performs flexible exposure adjustment such as underexposure correction, overexposure correction, local exposure adjustment, and exposure control. The flexibility is made possible by introducing a condition exposure map as input, which can be pre-defined or manually set. For (a) and (b), we use a pre-defined uniform exposure value for the exposure map. 	We can also assign non-uniform exposure values to different regions of an exposure map, as  shown in (c). Besides, multiple results of different exposure levels can be produced by assigning different exposure values to the exposure map, as shown in (d).  Our method adopts a zero-reference learning framework, thus it does not require paired or unpaired training data.}
	\vspace{-1.3em}
	\label{fig:example}
\end{figure*}

Despite the progress in underexposed image enhancement \cite{Guo2020CVPR,Guo2017,Jiang2019,survey,RUAS2021,Wang2019,Xu2020CVPR,LedNet2022ECCV,yan2025hvi}, most existing methods cannot handle diverse exposure properties. For example, underexposed image enhancement methods usually overexpose the normal-light region of a backlit image when they enhance the underexposed region.
Although there are several exposure correction methods \cite{Mahoud2021,DeepExposure,Lu2012,backlit2022,li2025osmamba}, they do not allow one to further adjust the exposure level.
Besides, these exposure correction methods commonly need heavy networks and have a slow inference speed. 

In this paper, we present an effective yet efficient method to cope with diverse exposure properties with a single model.
In contrast to previous studies, our method is more flexible in controllable exposure adjustment.
Our method is also appealing in its lightweight structure and fast inference speed. 
In particular, the network in our approach has only 3K trainable parameters.
By contrast, the recent exposure correction method \cite{Mahoud2021} needs 7M trainable parameters and the recent underexposed image enhancement method \cite{INN} needs 12M trainable parameters.
Furthermore, to process an image of 4K resolutions on a CPU, our method needs only 0.5s, while the runtime of most deep exposure correction models is unacceptable.
Such highly efficient and controllable exposure adjustment is achieved through \textit{novel curve distillation} and \textit{a new self-supervised spatial exposure control loss}. We provide an overview of these components as follows:

\noindent \textbf{Curve Distillation.}
Our method inherits the zero-reference learning and curve-based framework pioneered by Zero-DCE \cite{Guo2020CVPR}, an effective method for low-light image enhancement.
Different from the conventional Zero-DCE framework, we present unique advantages with the proposed curve distillation. 
(1) \textit{Our method significantly improves the inference speed of Zero-DCE}.
As illustrated in Figure \ref{fig:motivation}, we found that the output pixel value after the adjustment by applying a high-order curve can be approximated by the tangent line of the high-order curve near the point of tangency.
Moreover, the monotonicity relationships between neighboring pixels in the result adjusted by the high-order curve (neighboring pixels use the similar curve) can be inherited by the high-order curve's tangent line (neighboring pixels use the similar tangent linear), as shown in the zoomed-in box of Figure \ref{fig:motivation}.
Therefore, the time-consuming iteration operation in implementing the high-order curve can be replaced by a linear function that represents the high-order curve's tangent line.
(2) \textit{Our method reduces the model size of Zero-DCE}. 
Our curve distillation allows us to transfer compact knowledge from a large teacher network that accurately estimates the high-order curve's parameters.
By treating the result adjusted by the high-order curve as a supervisory signal, we can train a lightweight student network to estimate the tangent line of the high-order curve, achieving an efficient model.
It is worth noting that such distillation is effective due to the inherent approximating relationship of the curve and its tangent line near the point of tangency, the performance of which cannot be achieved by directly training a lightweight student network.

\noindent \textbf{Self-Supervised Spatial Exposure Control Loss.}
This newly proposed loss enables our method to perform controllable exposure adjustment beyond the underexposure enhancement as in conventional curve-based method \cite{Guo2020CVPR} and exposure correction as in recent method \cite{Mahoud2021}. 
This loss constrains the exposure levels of different spatial regions in a result close to the brightness distribution in an input condition exposure map with spatially variant random values in the training stage. Such a constrain helps the network in learning to exploit the exposure map as a condition for controllable exposure enhancement. 
Specifically, after training, our method can adjust a photo's exposure levels with the guidance of the input condition exposure map, which can be pre-defined or manually set in the inference stage.
As shown in Figure \ref{fig:example}, our method provides compelling results to correct underexposed and overexposed images and enables diverse and flexible exposure adjustment with a single model.

\noindent \textbf{Contributions.} We summarize our contribution as follows:
\begin{itemize} 
	\item We present curve distillation, a novel way to distill knowledge from large curve-based teacher network, achieving significant improvement in both the inference speed and model size. 
	\item We introduce a new self-supervised spatial exposure control loss to control the exposure levels of spatial regions of an image, endowing our method with the flexibility in controllable exposure adjustment.
	\item We propose the first zero-reference learning-based exposure adjustment network, which copes with diverse exposure properties with a single model and achieves state-of-the-art performance in real scenes. 
\end{itemize} 

Note that our work focuses on exposure correction, which aims to restore proper luminance in under- or over-exposed images. This is different from tone mapping, which compresses HDR content for display, and from contrast enhancement, which amplifies intensity differences to improve perceptual visibility. While exposure correction may appear related to low-light enhancement, we do not explicitly address noise removal, which is orthogonal to adjusting exposure.

\begin{figure*} [!t]
	\begin{center}
		\begin{tabular}{c@{ }}
			\includegraphics[width=0.85\linewidth]{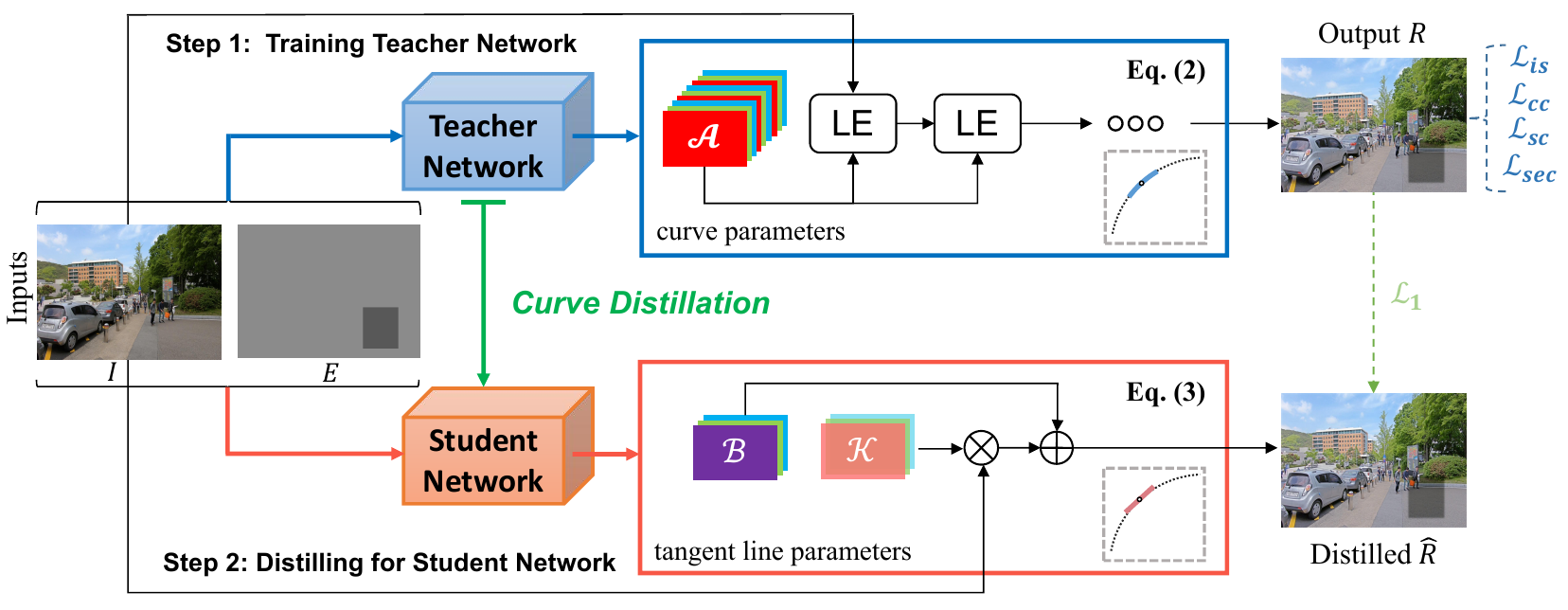}
		\end{tabular}
	\end{center}
	\vspace{-1.5em}
	\caption{\textbf{Framework Overview.} Our framework consists of two steps: Step 1: Training Teacher Network; Step 2: Distilling for Student Network. Given an image $I$ and a condition exposure map $E$ (its value is randomly set in the training stage and pre-defined or manually set in the inference stage), we feed  the two inputs to a teacher network to estimate the  parameters ($\mathcal{A}$) of quadratic curves  $\text{LE}$ that are iteratively applied for obtaining a high-order curve (Eq. \eqref{equ_2}). The high-order curve maps the input image $I$ to the output $R$. A set of non-reference losses is used to train the teacher network. After training the teacher network, we fix its weights and then distill it for a lightweight student network. The two inputs are simultaneously fed to the teacher network and the student network. Taking the high-order curve's output $R$ as supervision, the student network estimates the parameters (slope map: $\mathcal{K}$ and intercept map: $\mathcal{B}$) of the high-order curve's tangent line (Eq. \eqref{equ_3}). Then, the tangent line can map the input image $I$ to the distilled $\hat{R}$ that approximates the output $R$ of the high-order curve. \textit{After curve distillation, we only use the distilled student network in the inference stage.}}
	\vspace{-1em}
	\label{fig:framework}
\end{figure*}

\section{Related Work}
\label{sec:Related_Work}

\noindent
\textbf{Underexposure Enhancement.}  
A rich body of literature focuses on underexposed image enhancement.
Conventional methods \cite{Fu2016,Guo2017,Li2018,Ma2020,Wang2013} usually adopt Retinex model \cite{Retinex} to decompose an image into an illumination component and a reflectance component, where the illumination component is responsible for illuminance adjustment.
%
Recently, deep learning-based underexposed image enhancement methods \cite{Guo2020CVPR,Jiang2019,ZeroDCE++,RUAS2021,Lore2017,Chen2018,Xu2020CVPR,Yang2020CVPR,KinD++,KinD,INN,UTVNet2021,SCI2022,SNR2022,URetinexNet,survey,wu2023learning,cai2023retinexformer,yi2023diff,zhou2025low} have obtained increasing attention.
%
For example, Wu et al. \cite{wu2023learning} introduced the semantic priors to improve the performance of low-light image enhancement models, especially in color recovery.
%
Cai et al. \cite{cai2023retinexformer}  proposed a one-stage Retinex-based Transformer framework for low-light image enhancement. The framework can use illumination representations to direct the modeling
of non-local interactions of regions with different lighting
conditions.
%
Yi et al. \cite{yi2023diff} formulated the lowlight image enhancement problem into Retinex decomposition and conditional image generation, which integrates the advantages of the physical model and the generative network.
%
Different from most solutions that rely on paired data for supervised training, unsupervised learning-based methods \cite{Guo2020CVPR,Jiang2019,ZeroDCE++,RUAS2021,fu2023learning,liang2023iterative,lin2025aglldiff} have demonstrated better generalization capability in real scenes.
%
Among them, the zero-reference learning-based curve estimation method, Zero-DCE \cite{Guo2020CVPR,ZeroDCE++}, has shown encouraging results for practical applications.
Zero-DCE is independent of paired or unpaired data via a series of non-reference losses.
It formulates light enhancement as a task of image-specific curve estimation. 
In particular, quadratic curves are iteratively used to obtain a high-order curve, performing pixel-wise adjustment on the dynamic range of the input image.
Despite its promising performance, the time-consuming iterative operation constrains its inference speed.
Moreover, to  preserve the relationship among neighbouring pixels, Zero-DCE uses a large Unet-like network without feature scaling to estimate the quadratic curve's parameters. 
%
An accelerated and light version of Zero-DCE, called Zero-DCE++, was proposed by re-designing the network
structure, reformulating the curve formation, and controlling the sizes
of input image \cite{ZeroDCE++}.

\noindent
\textbf{Exposure Correction.}  
Apart from underexposure enhancement, there are several methods \cite{Mahoud2021,Exposure,DeepExposure,Lu2012,wang2023decoupling,yang2023learning} specially designed for coping with both underexposed and overexposed images. 
%
Based on the observation that, in the local regions covered
by convolution kernels, the feature response of low-/high-frequency can be decoupled by addition/difference operation,  Wang et al. \cite{wang2023decoupling} proposed to decouple the contrast enhancement and detail restoration within each convolution process for image exposure correction.
%
Different from traditional histogram-based methods, Afifi et al. \cite{Mahoud2021} proposed a promising coarse-to-fine exposure correction network, which progressively corrects the exposure errors by Laplacian pyramid decomposition and reconstruction. 
To perform supervised training, synthetic data with over-/under-exposed photos and the corresponding properly exposed photos are used in this method. 
In comparison to  Afifi et al~\cite{Mahoud2021}, our method does not require paired and unpaired training data, avoiding the risk of overfitting on specific datasets and thus could generalize better to diverse scenes. 
Different from the heavy reconstruction network (7M parameters) proposed in Afifi et al~\cite{Mahoud2021}, our model is lightweight (3K parameters), which is more promising for practical applications. 
Besides, our method is able to adjust the exposure levels of a photo, which cannot be achieved by previous methods.


\section{Our Approach}

\noindent
\textbf{Preliminary.}
Our approach is inspired by Zero-DCE \cite{Guo2020CVPR} that estimates pixel-wise and high-order curves for low-light image enhancement. 
Zero-DCE  formulates  light enhancement ($\text{LE}$)  as a quadratic curve:
\begin{equation}
	\label{equ_1}
	\text{LE}(I)=I(x)+\alpha I(x)(1-I(x)),
\end{equation}
where $I\in[0,1]$ is the input image, $x$ denotes the pixel coordinates, $\alpha\in[-1,1]$ adjusts the magnitude of the quadratic curve for dynamic range adjustment. 
To obtain powerful dynamic range adjustment, Zero-DCE iteratively applies the quadratic curve $n$ times to obtain a high-order curve:
\begin{equation}
	\label{equ_2}
	\text{LE}_{n}(I)=\text{LE}_{n-1}(I)+\mathcal{A}_{n} \text{LE}_{n-1}(I)(1-\text{LE}_{n-1}(I)),
\end{equation}
where $\mathcal{A}$ is a pixel-wise parameter map with the same size as the input image.  $\text{LE}_{n}$ is a $n$-th orders curve that adjusts the pixel value. The iteration number $n$ is set to eight. $R=\text{LE}_{n}(I)$ is the result adjusted by the high-order curve. For an image of three color channels, the high-order curve with different parameter maps is separately applied to each color channel. For eight iterations, 24 parameter maps (each set contains eight maps for one color channel) are estimated.
Besides, the good performance of Zero-DCE also benefits from the constrain of monotonicity
relationships between neighboring pixels, i.e., the neighbouring pixels in a local region can use a similar adjustment curve.

\subsection{Curve Distillation}
\label{curve}

In Figure \ref{fig:framework}, we present  \textbf{Cu}rve \textbf{Di}stillation (\textbf{CuDi}), a novel way to approximate the expensive high-order curve (Eq. \eqref{equ_2}) in Zero-DCE by estimating the high-order curve's tangent line: $\text{TL}(I)=kI(x)+b$,
where $k$ (i.e., slope) and $b$ (i.e., intercept) are the learnable  parameters that represent the tangent line of a high-order curve. 
Following the pixel-wise curve parameters in Zero-DCE, we also use the pixel-wise $k$ and $b$ (denoted as slope map $\mathcal{K}$  and intercept map $\mathcal{B}$) to approximate the performance of high-order curve: 
\begin{equation}
	\label{equ_3}
	\text{TL}(I)=\mathcal{K}I+\mathcal{B},
\end{equation}
where \text{TL}($I$) is the result adjusted by the tangent line, approaching the result of its corresponding high-order curve $\text{LE}_{n}(I)$. The  $\mathcal{K}$  and $\mathcal{B}$ have the same size as input image, and each one has three channels for adjusting three color channels. 

Our framework consists of two steps: Step 1: Training Teacher Network; Step 2: Distilling for Student Network. We detail the  curve distillation process as follows.

\noindent
\textbf{Step 1: Training Teacher Network.}
As shown in Figure \ref{fig:framework}, given an input image $I$ and a condition exposure map $E$, we first feed the two inputs to a large teacher network to estimate the parameters ($\mathcal{A}$) of quadratic curves $\text{LE}$  that are iteratively applied for obtaining a high-order curve, as defined in Eq. \eqref{equ_2}. 
The high-order curve maps the input image $I$ to the output $R$. 
The value of the condition exposure map  $E$ is set randomly in the training stage while it can be pre-defined or manually set in the inference stage for implementing controllable exposure adjustment. 
More details of the exposure control via exposure map condition are provided in Sec. \ref{exposuremap}.
For the teacher network, we do not use the same network structure as Zero-DCE. Instead, we use a larger Unet-like network without feature scaling that has 4.7M parameters as a baseline. The larger teacher network can achieve more accurate curve parameter estimation. 
Following the zero-reference learning framework of Zero-DCE, we also use non-reference losses to train the teacher network, detailed in Sec. \ref{loss}. Among these losses, we propose a new self-supervised spatial exposure control loss, in which the input exposure map $E$ as condition is fed to the teacher network for controlling the exposure level of the  result adjusted by the high-order curve. 

\begin{figure} [!t]
	\begin{center}
		\begin{tabular}{c@{ }}
			\includegraphics[width=0.9\linewidth]{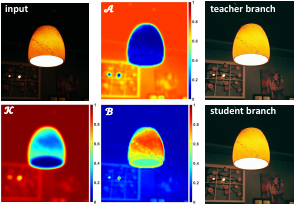}
		\end{tabular}
	\end{center}
	\vspace{-1em}
	\caption{\textbf{Results of curve distillation.} After curve distillation, the student branch (lightweight student network+tangent line) can achieve a similar result to the teacher branch (large teacher network+high-order curve). We normalize the values of parameter maps ($\mathcal{A}$) of high-order curve, and the slope map ($\mathcal{K}$) and intercept map ($\mathcal{B}$) of tangent line  to the range of [0,1] and average each set of parameter maps for visualization in heatmaps.}
	\vspace{-1.2em}
	\label{fig:curvemap}
\end{figure}

\noindent
\textbf{Step 2: Distilling for Student Network.}
After training the teacher network, we fix its weights. Given an input image $I$ and a condition exposure map  $E$, we simultaneously feed them to the teacher network and a student network. 
We take the results adjusted by the high-order curve as supervision information to train the student network that estimates the parameters (slope map: $\mathcal{K}$ and intercept map: $\mathcal{B}$) of the high-order curve's tangent line. 
Then, the tangent line (Eq. \eqref{equ_3}) can map the input image $I$ to the distilled $\hat{R}$ that approximates the output $R$ adjusted by the high-order curve.
After that, we only use the distilled student network in the inference stage.
As a baseline, we use an extremely lightweight network that has only 3K parameters as the student network, where we replace the convolutions with the depthwise separable convolutions \cite{MobileNets} for reducing computational resources. 
To further reduce the computational burden, we use downsampled image and exposure map as the input of the student network to estimate the slope map and intercept map of the linear function.
By default, we downsample an input by a factor of 4 to balance performance and computational cost.
The estimated slope and intercept maps are interpolated to the original resolution of input for linear mapping.
Such an interpolation operation does not affect the final result because the slop and intercept maps mainly contain low-frequency information, as the visualization shown in Figure \ref{fig:curvemap}.
The detailed network structures and parameters of the teacher network and student network are presented in Sec. \ref{sec:Experiments}.

\noindent
\textbf{Discussion.}
Compared with directly training a student network using non-reference losses, the slope and intercept maps of the tangent line function are easier to be estimated under the supervision of paired data (i.e., input and the result adjusted by the high-order curve).
This is because the results adjusted by the high-order curve capture the monotonicity relationships of neighboring pixels due to the curve formulation used in Zero-DCE.
Hence, using such results as the supervisory signal, the slope and intercept maps of the tangent line function, which are in a compact solution space constrained by monotonicity, can be learned more easily.
In Figure \ref{fig:curvemap}, the results are similar between the student branch and the teacher branch. Moreover,  like the parameter map ($\mathcal{A}$) of the high-order curve, the slope and intercept maps ($\mathcal{K}$, $\mathcal{B}$) of the high-order curve's tangent line also imply the brightness distribution of the input image. 
More discussions are presented in Sec. \ref{abl}.

\begin{figure*} [t]
	\begin{center}
		\begin{tabular}{c@{ }}
			\includegraphics[width=0.95\linewidth]{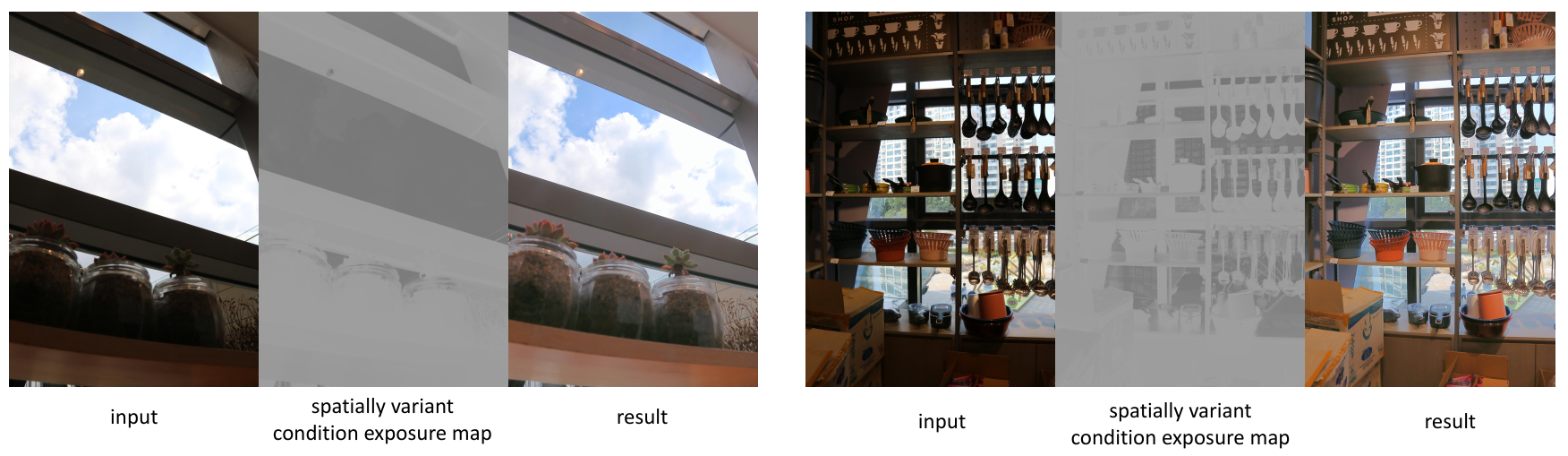}
		\end{tabular}
	\end{center}
	\vspace{-1em}
	\caption{\textbf{Results achieved by spatially variant condition exposure maps.} The spatially variant condition exposure maps are generated in our proposed way, i.e., we automatically assign the underexposed regions large exposure values (light grey) and well-exposed/overexposed regions small exposure values (dark grey).}
	\label{fig:map}
\end{figure*}

\subsection{Condition Exposure Map}
\label{exposuremap}
We use an input exposure map $E$ as condition to control the exposure level. 
The one-channel exposure map has the same size as the input image. 
During the training stage, we randomly select a region with an arbitrary shape from the exposure map, then assign different random values to the inside and outside of the region. The values range from 0.2 to 0.8.
With our proposed self-supervised spatial exposure control loss, we train the network such that it accepts the exposure map as guidance, producing a result with a similar brightness distribution to the input exposure map.
Hence, in the inference stage, we can flexibly control the global and local exposure of an image by setting the input exposure map that serves as a condition.

During inference, the value of the input exposure map can either be pre-defined or manually set according to different applications.
For instance, we can set a uniform exposure value, e.g., $E$=0.65 for underexposure correction and $E$=0.2 for overexposure correction. 
We can also set non-uniform exposure values to different regions of an exposure map, e.g., the underexposed region is assigned a large exposure value while the overexposed region is assigned a small exposure value.
One could apply different ways to generate a spatially variant condition exposure map.  We provide a simple way as follows. 
Specifically, we first obtain the luminance channel $L$ and its average value $L_{avg}$ of an input image, and then calculate the spatially variant condition exposure map by
$S+A\times \text{Norm}(L_{avg}-L)$, where $S$ is the base exposure value, $A$ is the adjustment amplitude, and $\text{Norm}$ operation normalizes its input to the range of [-1,1]. We empirically set $A$ to 0.15 and respectively set  $S$ to 0.55 and 0.25 for underexposed images and  overexposed images. In Figure \ref{fig:map}, we show two sets of results of our method using the spatially variant condition exposure maps. By using such a setting, we assign the underexposed regions large exposure values and well-exposed/overexposed regions small exposure values. 
Such a spatially variant setting can achieve satisfactory performance for elaborate exposure adjusting.
Besides, multiple results of different exposure levels can be produced using different exposure values.
%

\subsection{Loss Functions}
\label{loss}
To train the teacher network, we inherit the zero-reference learning framework of Zero-DCE, which is independent of paired or unpaired training data by non-reference losses.
Besides the color constancy loss $\mathcal{L}_{cc}$ that corrects the potential color deviations in the adjusted results, the illumination smoothness loss $\mathcal{L}_{is}$ that preserves the monotonicity relations between neighboring pixels, and the spatial consistency loss $\mathcal{L}_{sc}$ that preserves the spatial coherence proposed in Zero-DCE, we propose a new loss to implement more flexible and effective exposure adjustment, i.e., a self-supervised spatial exposure control loss $\mathcal{L}_{sec}$. We detail these losses below.
 
\noindent
\textbf{Self-Supervised Spatial Exposure Control Loss.}
To allow flexible control of the exposure levels of an image, we propose a self-supervised spatial exposure control loss $L_{sec}$.
This loss measures the distance of the average intensity value in the local regions of result and the input condition exposure map $E$. Thus, the exposure map as the input condition can control the exposure level of the final result. 
\begin{equation}
	\label{equ_5}
	\mathcal{L}_{sec}=\frac{1}{M}\sum\limits_{m=1}^M\| \text{Mean}(R_{m})-\text{Mean}(E_{m})\|_{1},
\end{equation}
where $M$ represents the number of non-overlapping local region $m$ with the size of 16$\times$16.
$\text{Mean}(R_{m})$ and $\text{Mean}(E_{m})$ are the average intensity values of $m$th local region of final result and input exposure map, respectively.

\noindent
\textbf{Spatial Consistency Loss.} 
To ensure spatial coherence between the input image and its adjusted version, we employ a spatial consistency loss,  which can be expressed as:
\begin{equation}
	\label{equ_4}
	\begin{aligned}
		\mathcal{L}_{sc}=\frac{1}{K}\sum\limits_{i=1}^K\sum\limits_{j\in\Omega(i)}(&\|
		\text{Mean}(R_{i})-
		\text{Mean}(R_{j})\|_{1}-\\
		&\|\text{Mean}(I_{i})-\text{Mean}(I_{j})\|_{1})^2,
	\end{aligned}
\end{equation}
where $K$ is the number of local region $i$, and $\Omega$($i$) is the four neighboring regions (top, down, left, right) centered at the region $i$. We denote $\text{Mean}(I_{i})$  as the $i$th local region's average intensity value. The local region size  is set to 4$\times$4. 

\noindent
\textbf{Color Constancy Loss.}
Following the Gray-World color constancy hypothesis~\cite{Buchsbaum1980} that color in each sensor channel averages to gray over the entire image, the color constancy loss can correct the potential color deviations in the final result and also build the relations among the three adjusted color channels.
The color constancy loss $L_{cc}$ can be expressed as:
\begin{equation}
	\label{equ_66}
	L_{cc}=\sum\nolimits_{\forall(p,q)\in\varepsilon}(\text{Mean}(R^{p})-\text{Mean}(R^{q}))^2, 
\end{equation}
where $\text{Mean}(R^{p})$ denotes the average intensity value of $p$ channel in the final result $R$, ($p$,$q$) $\in \varepsilon=\{(r,g),(r,b),(g,b)\}$ represents a pair of channels.

\noindent
\textbf{Illumination Smoothness Loss.} The illumination smoothness loss is proposed to preserve the monotonicity relations between neighboring pixels in each curve parameter map ($\mathcal{A}$). The illumination smoothness loss $L_{is}$ is defined as:
\begin{equation}
	\label{equ_77}
	L_{is}=\frac{1}{N}\sum\limits_{n=1}^N\sum\limits_{c\in\xi}(\|\nabla_{x}\mathcal{A}_{n}^{c}\|_{1}+\|\nabla_{y}\mathcal{A}_{n}^{c}\|_{1})^2,
\end{equation}
where $N$ is iteration number, $\xi=\{r,g,b\}$, $\nabla_{x}$ and $\nabla_{y}$ are the horizontal and vertical gradient operations, respectively.

At last, multi-term losses are used to optimize the teacher network. The final loss $\mathcal{L}_{tea}$ can be expressed as:
\begin{equation}
	\label{equ_7}
	\mathcal{L}_{tea}=\lambda_{sec}\mathcal{L}_{sec}+\lambda_{sc}\mathcal{L}_{sc}+\lambda_{cc}\mathcal{L}_{cc}+\lambda_{is}\mathcal{L}_{is},
\end{equation}
where $\lambda_{sec}$,  $\lambda_{sc}$, $\lambda_{cc}$, and $\lambda_{is}$ are the corresponding weights and are set to 10, 1, 5, and 200, respectively.

In the distillation process, we fix the weights of the teacher network and then simply use $\mathcal{L}_{stu}$=$\mathcal{L}_{1}$ loss to constrain the results adjusted by the high-order curve and its tangent line. 
\begin{equation}
	\label{equ_8}
	\mathcal{L}_{stu}=\|\text{TL}(I)-\text{LE}_n(I)\|_{1},
\end{equation}
where $\text{TL}(I)$=$\mathcal{K}I+\mathcal{B}$ represents the result adjusted by the tangent line, $\text{LE}_n(I)$ represents the result adjusted by the high-order curve.
In this way, we achieve the distilled student network that can estimate the optimal slope map $\mathcal{K}$  and intercept map $\mathcal{B}$ of the tangent line.

\section{Experiments}
\label{sec:Experiments}

\begin{figure*}[!t]
	\centering
	\includegraphics[width=0.7\linewidth]{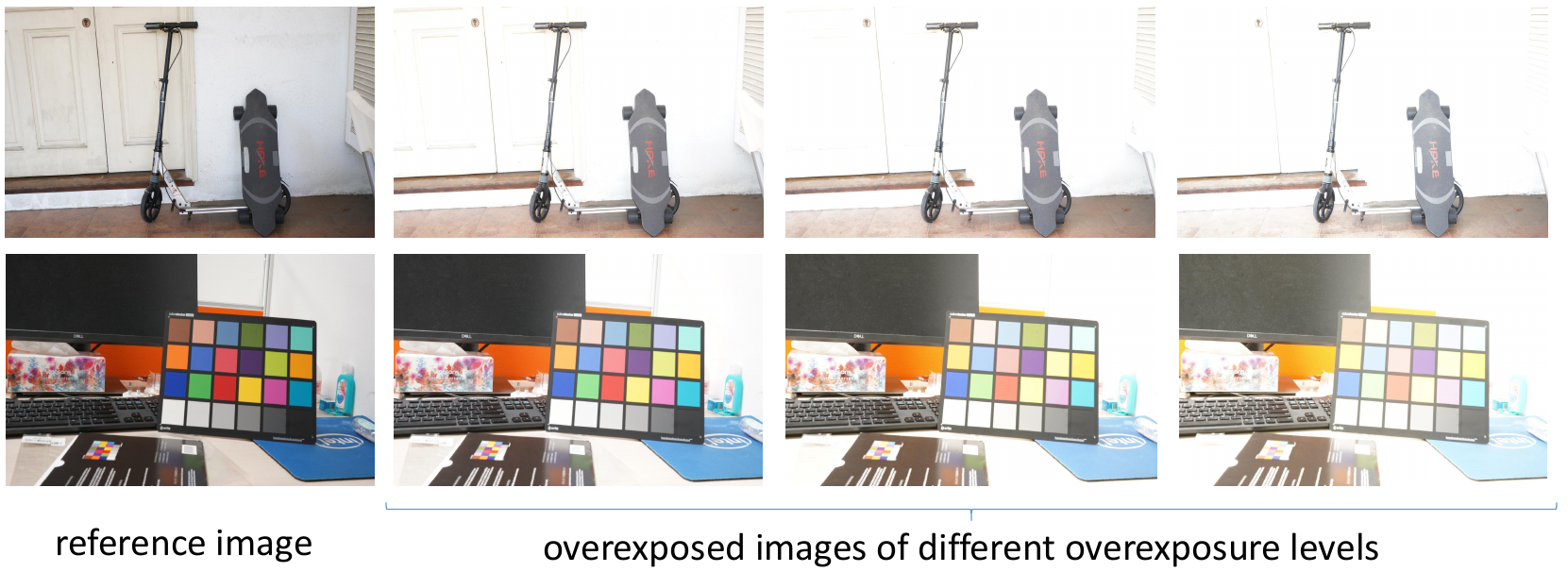}
	\caption{\textbf{Samples of the proposed OverExp-Pair dataset.} In this OverExp-Pair dataset, each well-exposed reference image has several overexposed images of different overexposure levels captured in the same scene.}
	\label{fig:dataset}
\end{figure*}

\noindent
\textbf{Implementation.} 
Our method is implemented with PyTorch on an NVIDIA Tesla V100 GPU.
We use an ADAM optimizer with default parameters for network optimization.
The fixed learning rates 0.0001 and 0.0005 are used for training the teacher network and the student network, respectively.
The filter weights of these two networks are initialized with the standard Gaussian function. 
A batch size of 8 is applied.
The size of training data is 256$\times$256.
Both the teacher network and the student network reach convergence at approximately 600 epochs, with the training process requiring six hours and three hours. Convergence was monitored via the total loss evaluated on a set of 100 randomly selected images.
In this part, we provide the detailed network structures and parameters of the teacher network and student network used in our method. 
The teacher network is a large Unet-like network that has 4.7M parameters while the student network is a lightweight  network that has only 3K parameters. 
In Table \ref{table:arch1}, we present the network structure and parameter comparisons between the teacher network and the student network. 
In addition, Table \ref{table:memory} details the peak GPU memory consumption (NVIDIA 3090 GPU) of our method across various image resolutions, while the hardware specifications of common edge devices are NVIDIA Jetson Nano: 4GB, Jetson Orin Nano: 8GB, and Raspberry Pi 5: 8–16GB. The results demonstrate that our method maintains a low memory footprint, ensuring high compatibility with standard edge computing platforms.

\begin{table}[h]
	\caption{\textbf{The peak GPU memory consumption of our method across various image resolutions.}}
		\vspace{-1em}
	\begin{center}
		\begin{tabular}{c|c|c|c|c}
			\hline
			 512$\times$512 &  2K &  4K & 6K & 8K  \\
			\hline
			\hline
			43.65MB  &184.80MB  &738.70MB
 &1662.56MB  &  2944.43MB \\
			\hline
		\end{tabular}
	\end{center}
	\label{table:memory}
		\vspace{-1em}
\end{table}

\begin{table}[!t]
	\vspace{-0.2cm}
	\caption{\textbf{Network structures and parameters of the teacher network and student network.} ``$Conv(3,64,3,1)$'' represents that $Conv(in\underline{~}channels=3,out\underline{~}channels=64,kernel\underline{~}size=3\times3,groups=1)$. ``$CSDN(in\underline{~}channels,out\underline{~}channels,kernel\underline{~}size,groups)$'' represents Depthwise Separable Convolutions.``Downsample: 4'' represents 4$\times$ downsampling operation while ``Upsample: 4'' represents 4$\times$ upsampling operation. $I$ is the input and $Y$ is the output. $LE_{8}(I)$ represents eight iterations of $LE$ function  in the high-order curve. $\mathcal{K}I+\mathcal{B}$ represents the tangent line.  }
	\centering
	\resizebox{9cm}{!}{
		\begin{tabular}{c|c|c}
			
			\bottomrule
			\textbf{Layer}  & \textbf{Teacher Network} & \textbf{Student Network}  \\
			\hline
			L1 & $ ReLU(Conv(4,32,3,1)) $   & Downsample: 4  \\
			&$ ReLU(Conv(32,32,3,1)) $  &$ CSDN(3,3,3,3) $  \\
			&$ ReLU(Conv(32,32,3,1)) $  & $ ReLU(CSDN(3,16,1,1))$ \\
			\hline
			L2 & $ ReLU(Conv(32,64,3,1)) $   & $ CSDN(16,16,3,16) $  \\
			&$ ReLU(Conv(64,64,3,1)) $  &$ ReLU(CSDN(16,16,1,1)) $ \\
			& $ ReLU(Conv(64,64,3,1)) $ &  \\
			\hline
			L3 & $ ReLU(Conv(64,128,3,1)) $   & $ CSDN(16,16,3,16) $  \\
			& $ ReLU(Conv(128,128,3,1)) $ &$ ReLU(CSDN(16,16,1,1)) $ \\
			& $ ReLU(Conv(128,128,3,1)) $ &  \\
			\hline
			L4 & $ ReLU(Conv(128,256,3,1)) $   & $ CSDN(16,16,3,16) $  \\
			& $ ReLU(Conv(256,256,3,1)) $ &$ ReLU(CSDN(16,16,1,1)) $ \\
			& $ ReLU(Conv(256,256,3,1)) $ &  \\
			\hline
			L5 & $ ReLU(Conv(256,256,3,1)) $   & $ CSDN(32,32,3,32) $  \\
			& $ ReLU(Conv(256,256,3,1)) $ &$ ReLU(CSDN(32,16,1,1)) $ \\
			& $ ReLU(Conv(256,256,3,1)) $ &  \\
			\hline
			L6 & $ ReLU(Conv(384,128,3,1)) $   & $ CSDN(32,32,3,32) $  \\
			& $ ReLU(Conv(128,128,3,1)) $ &$ ReLU(CSDN(32,16,1,1)) $ \\
			& $ ReLU(Conv(128,128,3,1)) $ &  \\
			\hline
			L7 & $ ReLU(Conv(192,64,3,1)) $    & $ CSDN(32,32,3,32) $  \\
			& $ ReLU(Conv(64,64,3,1)) $ & $ CSDN(32,6,1,1) $ \\
			& $ ReLU(Conv(64,64,3,1)) $ & Upsample: 4 \\
			\hline
			L8 & $ ReLU(Conv(96,32,3,1)) $   &   \\
			& $ ReLU(Conv(32,32,3,1)) $ &  \\
			& $ ReLU(Conv(32,32,3,1)) $ & \\
			\hline
			L9 & $ Tanh(Conv(32,24,3,1)) $   &   \\
			\hline
			Function & $Y=LE_8(I) $   &$Y=\mathcal{K}I+\mathcal{B}$ \\
			\hline
			\toprule
		\end{tabular}
	}
	\label{table:arch1}
\end{table}

\noindent
\textbf{Training Data.} 
For training our method, we choose 1,000 normal brightness images from REDS dataset \cite{REDS}, which was originally used for image deblurring and super-resolution. Our method is insensitive to the datasets. 
One can use other images of normal brightness to train our method.
Note that no paired or unpaired data are needed to train our zero-reference learning framework. 
To verify the generalization capability of our method, we conduct experiments on diverse images taken in real-world scenes.
We do not use synthetic data and focus only on  8-bit color images.

\noindent
\textbf{Testing Data.} 
For underexposure correction experiments, we use the paired data of the VE-LOL-L-Cap dataset \cite{ijcvsurvey}, in which each captured well-exposed image has its underexposed versions with different underexposure levels. However, some paired data of the VE-LOL-L-Cap dataset exhibit significant misalignment. We discard these misaligned data, obtaining  219 pairs of data.
Besides, following previous works \cite{Mahoud2021,Guo2020CVPR,ZeroDCE++}, we perform experiments on some underexposed image datasets, including NPE \cite{Wang2013}, LIME\cite{Guo2017}, MEF \cite{Makede15}, DICM \cite{Lee12}, and VV \cite{VV}. The testing set contains diverse underexposed images, denoted as UndExp-Web.

For overexposure correction experiments, we contribute a real dataset, called OverExp-Pair, which contains 180 pairs of well-exposed/overexposed images. Each well-exposed image has several overexposed versions with different overexposure levels. 
Specifically, a Sony $\alpha7$ \uppercase\expandafter{\romannumeral3} camera is first mounted on a tripod. 
Then, we capture an image with a small ISO $\in[100,500]$ in a light sufficient scene (indoor or outdoor).
After that, we prolong the exposure time to obtain overexposed images of the same scene. 
Despite the camera being mounted on a tripod, subtle misalignment may exist between the paired data.
To solve this issue, we adopt an image alignment algorithm (i.e., ECC \cite{ECC}) to align the overexposed image and its ground truth. 
We show several samples of our OverExp-Pair dataset in Figure \ref{fig:dataset}. 
Such a real overexposure correction dataset is scarce and the collection of such a dataset is non-trivial.

In addition, we also collect 70 diverse overexposed images from Flickr and websites for comparisons, denoted as  OverExp-Flickr dataset. 
Although our method can handle underexposed and overexposed images like previous methods, it has broader applications (examples shown in Figure \ref{fig:example}), thanks to its efficient and controllable nature. 
Since there is no baseline of this kind, we show more examples of broader applications of our method in Sec. \ref{subsec: diverse} and our project page.
The datasets, results, and code of this work will be released.

\noindent
\textbf{Runtime Computation.} 
We compare the runtime of different versions of our method and the compared methods. The runtime of the same network can be affected by factors such as device configuration, usage age, image resolution, and even temperature. For ultra-lightweight networks, these factors become even more significant, making it impossible to convert the runtime using a simple scaling relationship.  When testing ultra-lightweight networks, however, idle waiting of the GPU can lead to inaccurate runtime measurements. To unify the testing standard, instead of measuring runtime by repeatedly processing a single image, recording only the model execution time, and then averaging the results (after sufficient warm-up), we evaluate Zero-DCE, CuDi, and Zero-DCE++ on the same device (an NVIDIA Quadro RTX 8000 GPU or Intel(R) Xeon(R) CPU@2.60GHz) by generating random tensors (to shorten the time for reading images and reduce GPU idle time) and measuring execution time accordingly, as presented in Tables \ref{table:ComplexTS},\ref{table:ComplexTS2}, and \ref{table:zero-dce++}. For convenience, we define the resolutions of 2K, 4K, 6K, 8K, and 16K as (2048$\times$1080$\times$3), (4096$\times$2160$\times$3), (6144$\times$3240$\times$3), (7680$\times$4042$\times$3), and (15360$\times$8084$\times$3), respectively.

\subsection{Ablation Study}
\label{abl}
We conduct ablation studies to investigate the necessity and effectiveness of curve distillation proposed in our method.
Unlike supervised methods, our zero-reference learning framework works only when all mentioned non-reference losses are synergistic. The effect of removing individual losses has been studied in our prior work \cite{Guo2020CVPR}. To avoid redundancy, we do not repeat these experiments in this paper. 
Unless otherwise stated, all training settings remain unchanged as the implementation.

\begin{figure} [!t]
	\begin{center}
		\begin{tabular}{c@{ }}
			\includegraphics[width=1\linewidth]{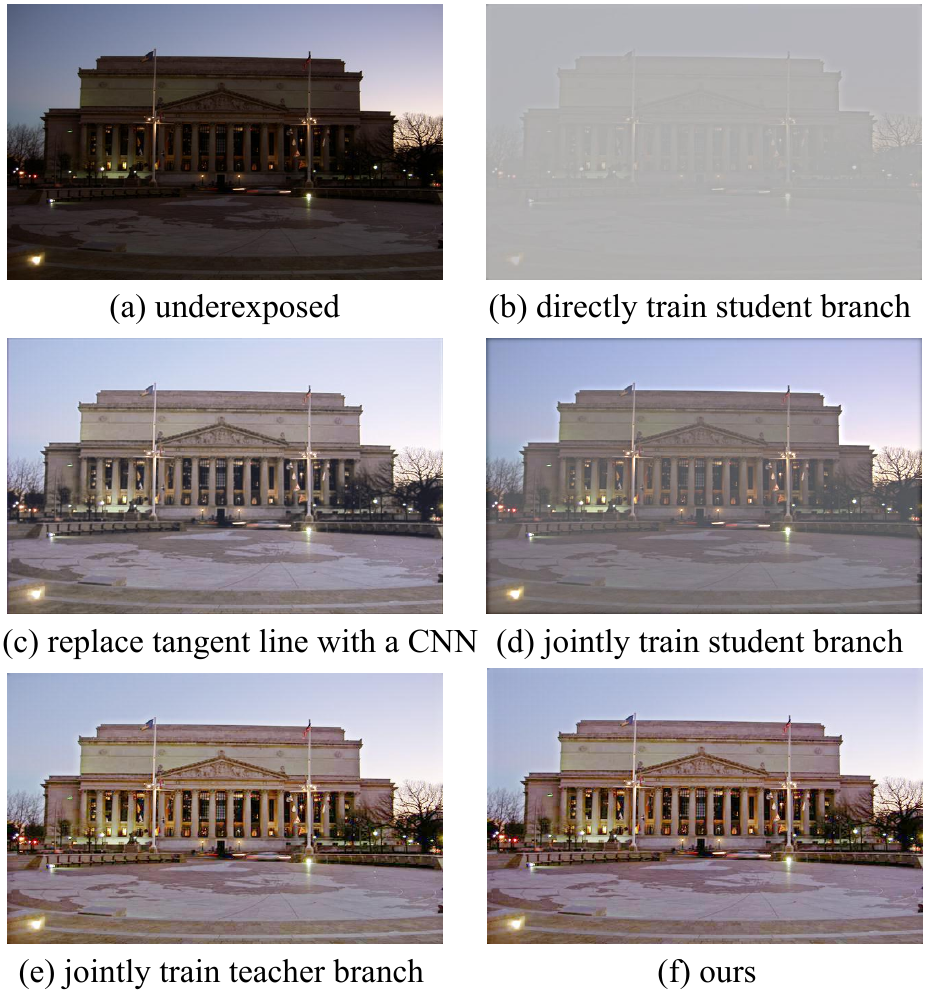}
		\end{tabular}
	\end{center}
    \vspace{-1em}
	\caption{\textbf{Visual comparisons of ablated curve distillation.} We set a uniform exposure value of 0.6 to the input condition exposure map, which is omitted for brevity.}
	\label{fig:dirtrain}
\end{figure}

\noindent
\textbf{Necessity of Curve Distillation.} 
To demonstrate the necessity of curve distillation, we directly train the student branch (student network+tangent line) using the same losses (including the self-supervised spatial exposure control loss, spatial consistency loss, color constancy loss, and illumination smoothness loss) and training data as the teacher branch (teacher network+high-order curve). Additionally, we add the illumination smoothness loss to tangent line parameters.
%
We also show the necessity of estimating the tangent line of a high-order curve by replacing the tangent line with a CNN with three convolutional layers where each layer has 64 kernels of a size of 3$\times$3 (student network+CNN) in the curve distillation process. 
Besides, we jointly train the teacher network and the student network, instead of training the teacher first and then the student network. The comparison results are shown in Figure \ref{fig:dirtrain}.

\begin{table*}[!t]
	\caption{\textbf{Comparisons of FLOPs, trainable parameters, and runtime between the teacher branch and the student branch for processing an image with a resolution of 2K, 4K, or 6K.}  The bold numbers denote the best performer in each case. ``-'' indicates that the result is unavailable due to the constraint of GPU memory.}
	\vspace{-1em}
	\begin{center}
		\begin{tabular}{c|c|c|c||c|c|c||c|c|c||c}
			\hline
			\multirow{2}{*}{Curve Distillation} & \multicolumn{3}{c||}{GPU$\downarrow$} & \multicolumn{3}{c||}{CPU$\downarrow$} & \multicolumn{3}{c||}{FLOPs$\downarrow$} & \multirow{2}{*}{parameters$\downarrow$}\\
			\cline{2-10}
			\cline{2-10}
			& 2K & 4K & 6K & 2K & 4K & 6K & 2K & 4K & 6K & \\
			\hline
			\hline
			Teacher Branch  & 1.6s & - & - &16.9s &64.9s &160.4s & 10.4T &41.6T & 93.6T & 4.7M  \\
			Student Branch & \textbf{4.9ms} & \textbf{23.9ms} & \textbf{46.1ms} & \textbf{0.1s} &\textbf{0.6s} & \textbf{1.3s} & \textbf{0.5G} &\textbf{2.1G} & \textbf{4.8G} & \textbf{3K}\\
			\hline
		\end{tabular}
	\end{center}
	\label{table:ComplexTS}
\end{table*}

\begin{table}[!t]
	\caption{\textbf{Comparisons of FLOPs and runtime between  the iterative operation (i.e., $LE_8(I)$) and the linear operation (i.e., $\mathcal{K}I+\mathcal{B}$) for processing images of different resolutions.}  The bold numbers denote the best performer in each case. In this table, we only include the computations of the iterative operation and the linear operation,  ignoring the computations of the rest modules.}
	\vspace{-1.5em}
	\begin{center}
		\begin{tabular}{c|c|c||c|c||c|c}
			\hline
			\multirow{2}{*}{Operations} & \multicolumn{2}{c||}{GPU$\downarrow$} & \multicolumn{2}{c||}{CPU$\downarrow$} & \multicolumn{2}{c}{FLOPs$\downarrow$} \\
			\cline{2-7}
			\cline{2-7}
			& 8K & 16K & 8K & 16K & 8K & 16K \\
			\hline
			\hline
			 $LE_8(I)$ & 58.1ms & 0.2s &1.7s &6.1s & 3.0G & 11.9G\\
			 $\mathcal{K}I+\mathcal{B}$ & \textbf{3.6ms}  & \textbf{0.01s} &\textbf{0.1s} &\textbf{0.4s} & \textbf{0.2G} & \textbf{0.7G}\\
			\hline
		\end{tabular}
	\end{center}
	\label{table:ComplexTS2}
\end{table}

As shown in Figure \ref{fig:dirtrain}(b), without curve distillation, the directly trained student branch produces visually unpleasing results.
This is mainly because the solution space of exposure adjustment is too complex to be directly learned by a simple linear function.
In addition, as shown in Figure \ref{fig:dirtrain}(c), using a CNN to replace the tangent line cannot achieve satisfactory results either.
%
Figures \ref{fig:dirtrain}(d) and (e) show that joint training of the teacher and student networks does not produce optimal results for the student network, even though the teacher network can be effectively trained. This is because simultaneously optimizing the teacher while distilling knowledge to the student is inherently challenging.
%
In Figure \ref{fig:dirtrain}(f), our proposed curve distillation successfully corrects the underexposed image.  

\begin{figure} [h]
	\begin{center}
		\begin{tabular}{c@{ }}
                \includegraphics[width=1\linewidth]{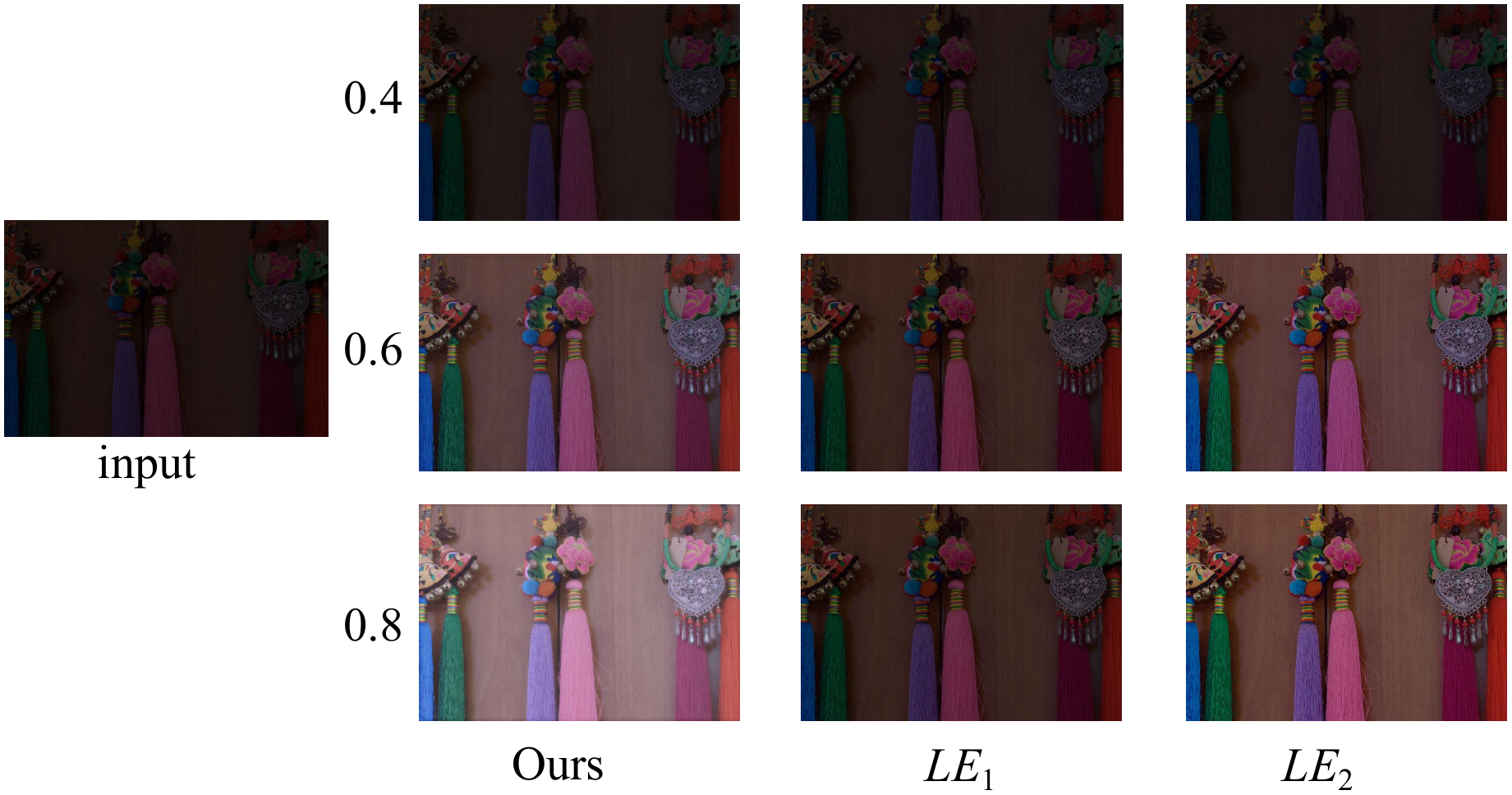}
                \vspace{0.1cm}
		\end{tabular}
	\end{center}
	\vspace{-2em}
	\caption{\textbf{Visual comparisons between our method (distilling $LE_8$ into its tangent line approximation) and  $LE_1$/$LE_2$ distilling. The numerical values labeled on the image denote the specific exposure settings used as conditions.}}
	\label{fig: LE1-2}
\end{figure}

To demonstrate the necessity of approximating $LE_8$ with its tangent-line formulation, we further conducted experiments using only one and two enhancement iterations, denoted as $LE_1$ and $LE_2$, respectively. In Fig. \ref{fig: LE1-2}, both $LE_1$ and $LE_2$ fail to produce satisfactory lightness adjustment, particularly under challenging exposure conditions. The limited number of iterations results in insufficient enhancement strength, making the model incapable of adjusting the illumination according to the target exposure values. In contrast, the proposed tangent-line approximation of $LE_8$ can effectively perform exposure adjustment while preserving the desired enhancement behavior.

It is worth noting that replacing $LE_8(I)$ with low-iteration variants such as $LE_1(I)$ or $LE_2(I)$ may appear to provide computational efficiency comparable to the proposed linear approximation. However, these low-iteration formulations fundamentally suffer from limited expressive capability. Due to the monotonicity constraint of the iterative mapping, the achievable slope under one or two iterations has an intrinsic upper bound, which restricts the maximum enhancement gain that can be produced in extremely dark or bright regions. Therefore, the performance degradation is not caused by insufficient optimization or distillation quality, but rather by the structural limitation of the low-iteration iterative formulation itself.

Therefore, although $LE_1$/$LE_2$ and the tangent-line approximation have similar computational complexity, they differ substantially in enhancement capability. The proposed tangent approximation preserves the monotonic property of the original formulation while enabling sufficiently large enhancement slopes within a single-step computation, making it both computationally efficient and practically necessary.

To further show the advantage of curve distillation, we compare the FLOPs, trainable parameters, and runtime between the teacher branch and the student branch in Table \ref{table:ComplexTS}. 
The results suggest that curve distillation can significantly speed up the inference time and reduce the model size.
Furthermore, we  compare the FLOPs and runtime of  the iterative operation (i.e., $LE_8(I)$) and the linear operation (i.e., $\mathcal{K}I+\mathcal{B}$),  ignoring the computations of the rest modules such as the parameter estimation networks. The comparison results are shown in Table \ref{table:ComplexTS2}. As presented, the linear operation needs less computational resources and is significantly faster than the iterative operation, especially for high-resolution images.

\noindent
\textbf{Effectiveness of Curve Distillation.} 
To show how similar the results adjusted by the teacher branch and the results distilled by the student branch are, we randomly choose 100 images (50 overexposed images and 50 underexposed images)  to compute the quantitative similarity of the outputs of the teacher branch and the distilled student branch.
We use MSE, SSIM \cite{SSIM}, and  Pearson Correlation Coefficient (PCC) \cite{PCC} as the evaluation metrics. PCC value 1  represents the highest correlation.
The results are shown in Table \ref{table:similarity}, in which the results of the teacher branch are treated as ground truths for measuring the quantitative similarity. 
Similar quantitative results also imply the effectiveness of curve distillation.

\begin{table}[!t]
	\caption{\textbf{Comparisons of MSE, SSIM, and PCC between the results of the teacher branch and the student branch.} To show the effectiveness of curve distillation, we treat the results of the teacher branch as ground truths to measure the  similarity.}
		\vspace{-1em}
	\begin{center}
		\begin{tabular}{c|c|c|c}
			\hline
			Curve Distillation & MSE$\downarrow$ &  SSIM$\uparrow$ &  PCC$\uparrow$ \\
			\hline
			\hline
			Teacher Branch &0  &1  &1\\
			Student Branch &0.092  &0.952   &0.998\\
			\hline
		\end{tabular}
	\end{center}
	\label{table:similarity}
\end{table}

\noindent
\textbf{Advantage of CuDi over Zero-DCE++.} 
Zero-DCE \cite{Guo2020CVPR} has a more efficient variant, called Zero-DCE++ \cite{ZeroDCE++}.  
Compared to Zero-DCE++, our CuDi introduces conditional input for exposure control, which requires additional parameters to understand these conditions and influence the results accordingly. This allows CuDi to effectively handle both overexposure and underexposure scenarios by appropriately setting the exposure conditions. In contrast, Zero-DCE++ is primarily designed for low-light image enhancement. While it also has the potential to address overexposure and underexposure, it requires careful adjustment of the proportion of overexposed and underexposed data in its training dataset, which often leads to suboptimal results due to the unpredictability of ideal exposure settings. Moreover, even with a well-balanced dataset, the absence of CuDi’s conditional input and exposure-control loss functions makes it challenging for Zero-DCE++ to adjust local exposure.

\begin{table}[!t]
	\caption{\textbf{Comparison of runtime (ms) between Zero-DCE++ and CuDi.} The bold numbers denote the best performer in each case.}
	\vspace{-1em}
	\begin{center}
		\begin{tabular}{c|c|c|c|c}
			\hline
			Resolutions & $512^{2}$ & $1024^{2}$ & $2048^{2}$ & $4096^{2}$ \\
			\hline
			\hline
			Zero-DCE++    &1.31   & 2.24  & 8.59  & 34.37\\
			CuDi   & \textbf{0.91}   & \textbf{0.93} & \textbf{3.68} & \textbf{14.77}\\
			\hline
		\end{tabular}
	\end{center}
	\label{table:zero-dce++}
         \vspace{-1em}

\end{table}

In addition, the computational efficiency of CuDi is improved when compared with Zero-DCE++, especially for the high-resolution input image, as shown in Table \ref{table:zero-dce++}. 
Moreover, CuDi is well-suited for edge computing on chips, as the linear function involves only Multiply-Add operations, enabling acceleration with Hi-chip and specific acceleration conditions. For instance, the processing speed of a  640$\times$480 image on the Hi3516DV300 device has improved from 120 ms per image to 17 ms per image when compared with Zero-DCE++.
At last, as demonstrated in previous works \cite{ZeroDCE++}, the visual performance of Zero-DCE and Zero-DCE++ is nearly identical. Therefore, considering the space constraints, we did not include the visual results of Zero-DCE++ again in the manuscript.

\begin{figure} [h]
	\begin{center}
		\begin{tabular}{c@{ }}
                \includegraphics[width=1\linewidth]{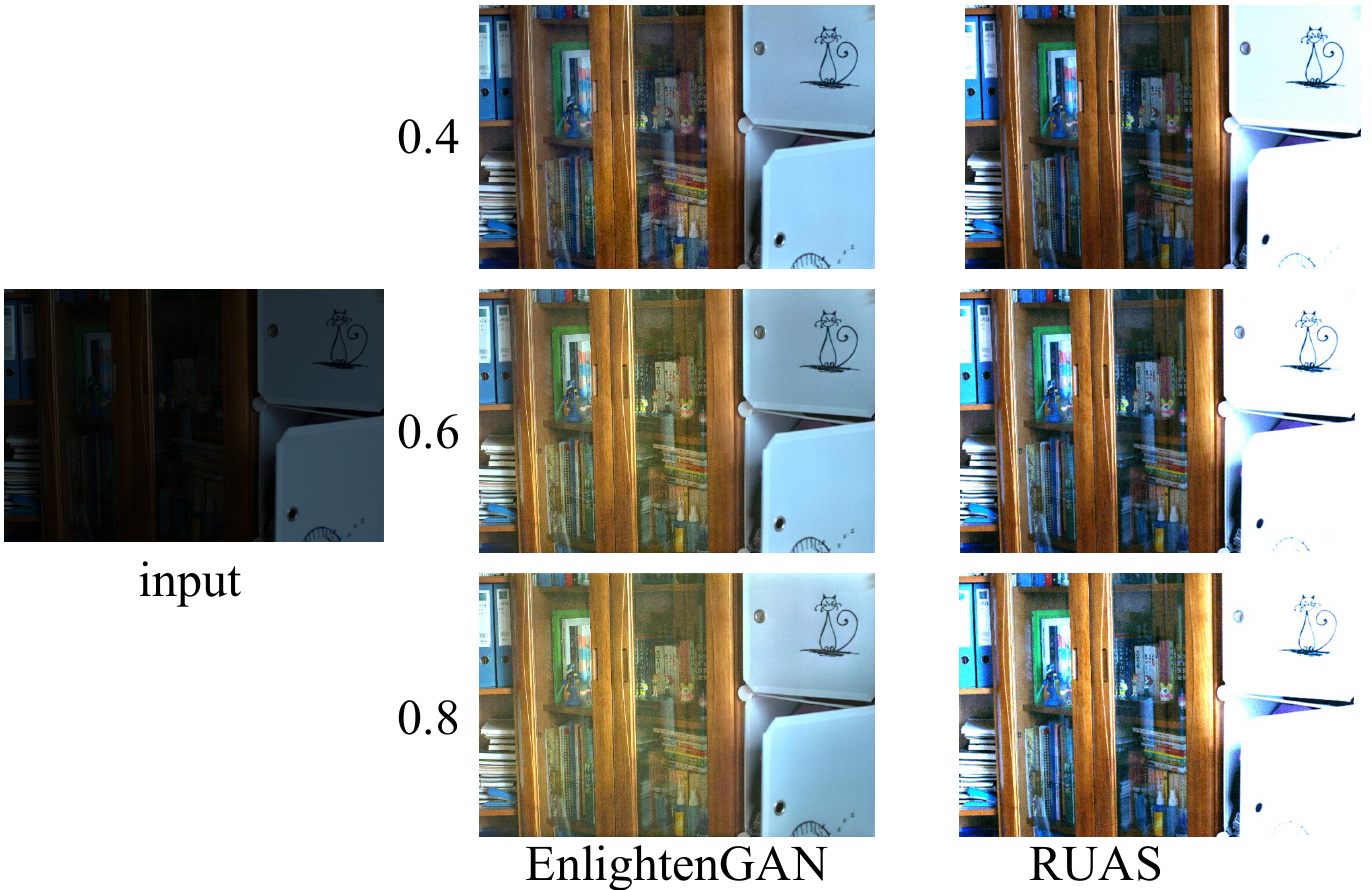}
                \vspace{0.1cm}
		\end{tabular}
	\end{center}
	\vspace{-1em}
	\caption{\textbf{The results of EnlightenGAN \cite{Jiang2019}  and RUAS \cite{RUAS2021} training with our controllable exposure. The numerical values labeled on the image denote the specific exposure settings used as conditions.}}
	\label{fig: exposure_loss}
\end{figure}

\noindent
\textbf{Effectiveness of Controllable Exposure.}  
To show that the proposed exposure map, exposure control loss, and proposed training augmentations can be applied to other exposure correction methods, we have applied them to EnlightenGAN \cite{Jiang2019}  and RUAS \cite{RUAS2021}. As shown in Fig. \ref{fig: exposure_loss}, the proposed controllable exposure is also effective in other exposure correction methods. 

\begin{figure*}[!t]
	\centering
	\includegraphics[width=0.95\linewidth]{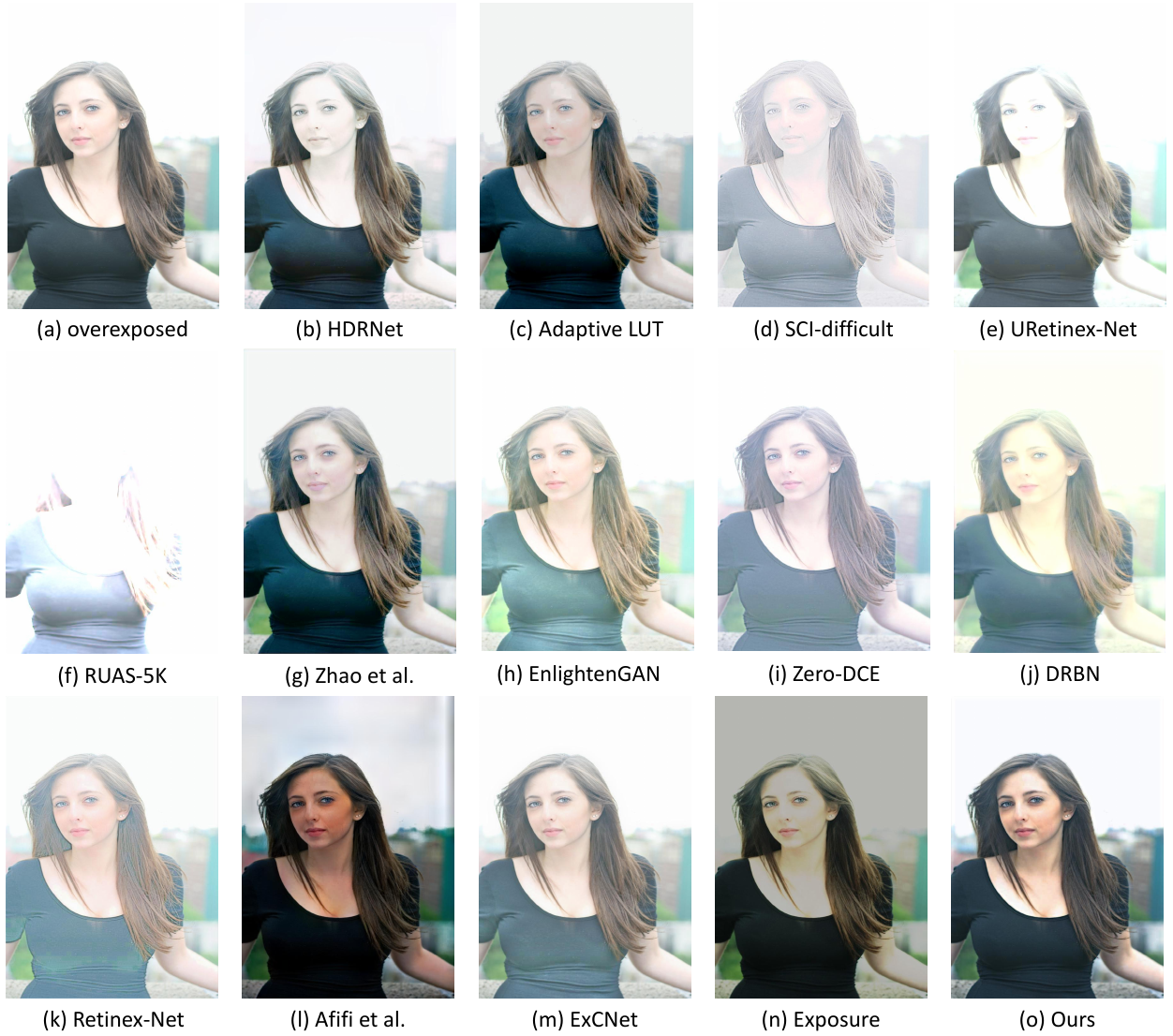}
	\caption{\textbf{Visual comparisons on an overexposed image sampled from OverExp-Flickr dataset.}}
	\label{fig:overexposed_unpair1}
	\vspace{-1em}
\end{figure*}

\begin{figure*}[!t]
	\centering
	\includegraphics[width=0.95\linewidth]{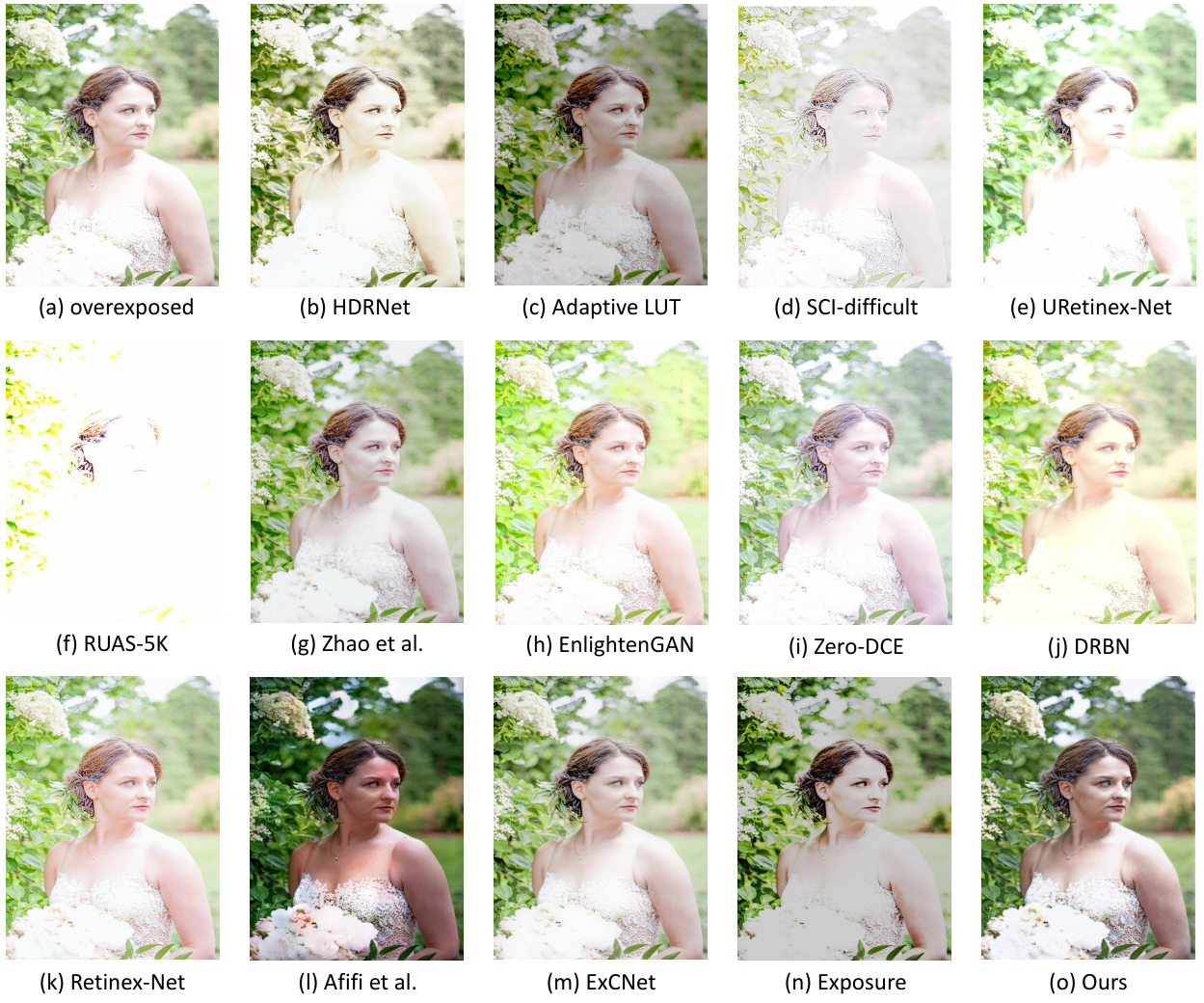}
	\caption{\textbf{Visual comparisons on an overexposed image sampled from OverExp-Flickr dataset.}}
	\label{fig:overexposed_unpair2}
\end{figure*}

\begin{figure*}[!t]
	\centering
	\includegraphics[width=0.95\linewidth]{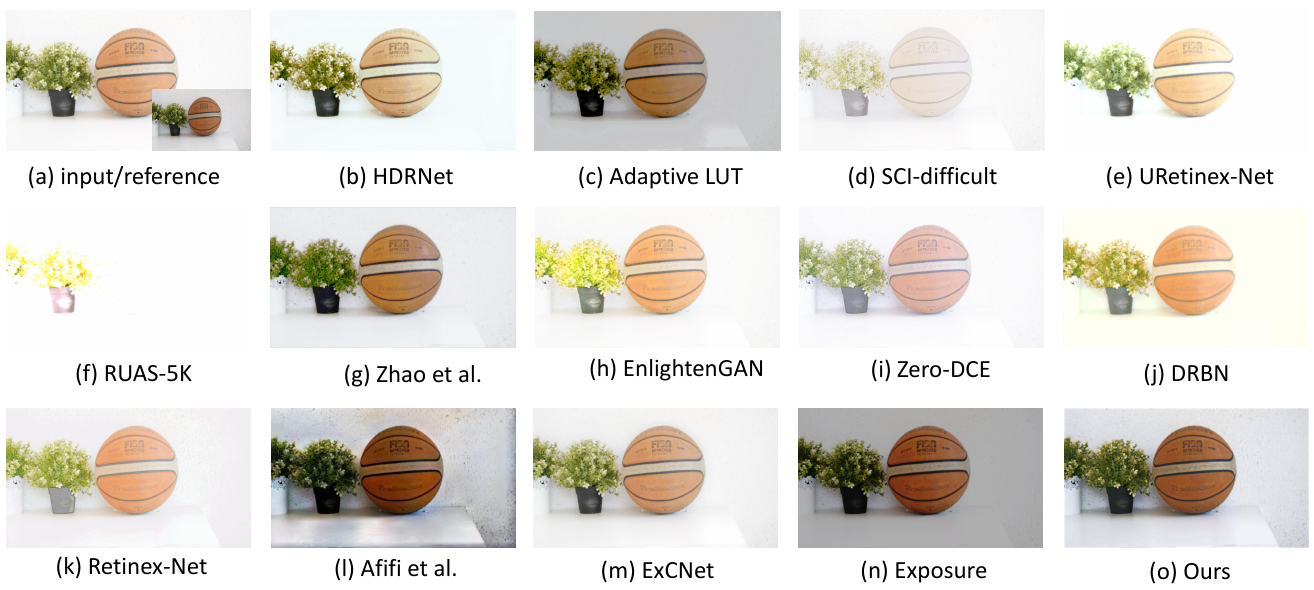}
	\caption{\textbf{Visual comparisons on an overexposed image sampled from OverExp-Pair dataset.}}
	\label{fig:overexposed_pair}
	\vspace{-1em}
\end{figure*}

\begin{figure*}[!t]
	\centering
	\includegraphics[width=0.95\linewidth]{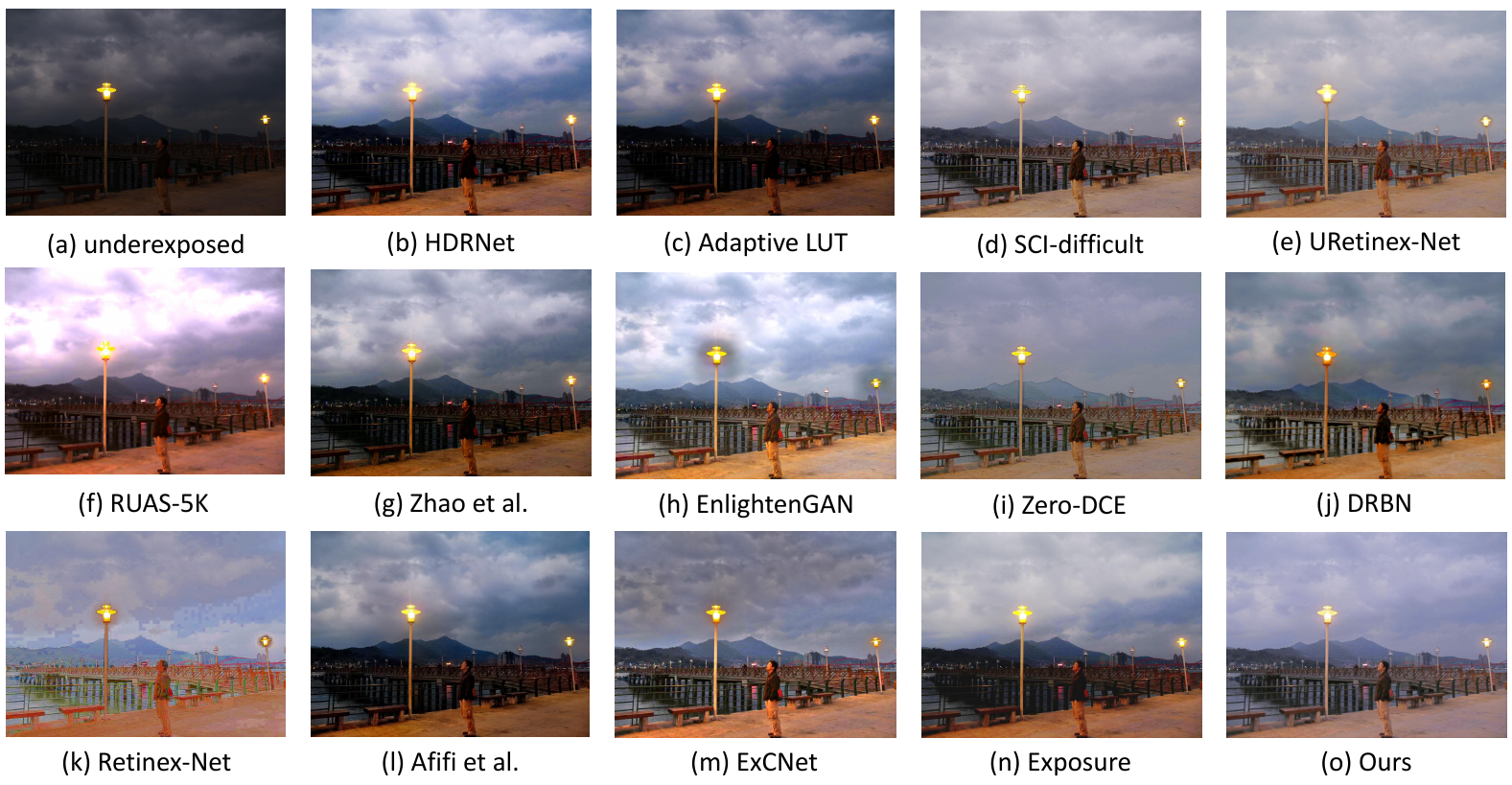}
	\caption{\textbf{Visual comparisons on an underexposed image sampled from UndExp-Web dataset.}}
	\label{fig:underexposed_unpair}
		\vspace{-1em}
\end{figure*} 

\begin{figure*}[!t]
	\centering
	\includegraphics[width=0.95\linewidth]{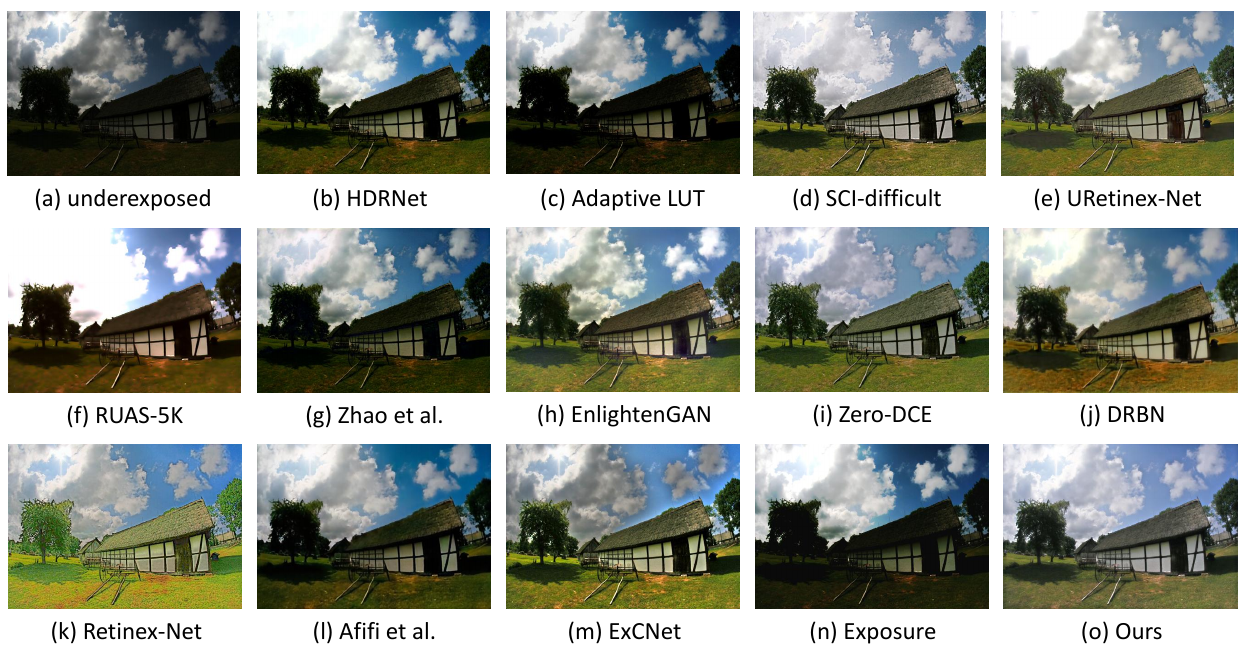}
	\caption{\textbf{Visual comparisons on an underexposed image sampled from UndExp-Web dataset.}}
	\label{fig:underexposed_unpair2}
		\vspace{-1em}
\end{figure*}

\subsection{Comparisons with State-of-the-art Methods}
\label{subsec: comparison}
We compare our method with 13 methods (17 models), including eight underexposure enhancement methods (SCI \cite{SCI2022}, URetinex-Net \cite{URetinexNet}, RUAS \cite{RUAS2021}, Zhao et al. \cite{INN}, EnlightenGAN \cite{Jiang2019}, Zero-DCE~\cite{Guo2020CVPR}, DRBN~\cite{Yang2020CVPR}, Retinex-Net \cite{Chen2018}), three exposure correction methods (Afifi et al. \cite{Mahoud2021}, ExCNet \cite{ExCNet}, Exposure \cite{Exposure}, 
and two efficient photo enhancement methods (HDRNet \cite{gharbi2017deep}, Adaptive LUT \cite{zeng2020learning}.
%
We use their pre-trained models and include two models (RUAS-5K and RUAS-LOL) of RUAS \cite{RUAS2021} and three models (SCI-easy, SCI-medium, and SCI-difficult) of SCI \cite{SCI2022} for comparison. 
Among these methods,  SCI \cite{SCI2022}, RUAS \cite{RUAS2021}, Zero-DCE~\cite{Guo2020CVPR}, ExCNet \cite{ExCNet}, and our method are zero-reference learning-based methods while the rest methods need supervised training.
%
We do not show the visual results of RUAS-LOL, SCI-easy, and SCI-medium due to their limited robustness.
The compared methods used the original code, and we did not make any modifications, including data pre-processing.

Since our method is able to infer multiple results rapidly by setting the condition exposure map with different exposure values, we provide two versions of our results. 
`Ours-auto': we fix the exposure value of the exposure map for all testing images. We use a uniform exposure value of 0.65 for underexposed images while it is set to 0.2 for overexposed images.
`Ours-manu': we manually set the exposure value of the exposure map for each testing image.
We only produce results using three fixed exposure values (0.55, 0.65, and 0.75 for underexposed images; 0.2, 0.25, and 0.3 for overexposed images) and subjectively choose the most visually pleasing result as the final result.
We do not have access to any reference image when we choose the final result. %
We use the results of ``Ours-auto'' for visual comparisons.

\begin{table*}[!t]
	\caption{\textbf{Quantitative comparisons on overexposed (OverExp-Pair, OverExp-Flickr) datasets and underexposed (VE-LOL-L-Cap, UndExp-Web) datasets.}  We denote methods designed for exposure correction in \textit{italic}. The best result is in \textcolor{red}{red} whereas the second and third  are in \textcolor{blue}{blue} and  \textcolor{brown}{brown} under each case, respectively. ``-'' indicates that the result is unavailable due to the Nan returned by the NIQE metric. $\dagger$ and $\star$ indicate the unsupervised method and zero-reference learning-based method, respectively. Other methods are supervised methods.}
		\vspace{-1em}
	\begin{center}
		\resizebox{18cm}{!}{
			\begin{tabular}{c|c|c|c|c|c|c||c|c|c|c|c|c}
				\hline 
				& \multicolumn{6}{c||}{Overexposure Correction} & \multicolumn{6}{c}{Underexposure Correction} \\
				\cline{2-13}
				\multirow{2}{*}{Methods} & \multicolumn{3}{c|}{OverExp-Pair} &\multicolumn{3}{c||}{OverExp-Flickr} & \multicolumn{3}{c|}{VE-LOL-L-Cap} &\multicolumn{3}{c}{UndExp-Web}  \\
				\cline{2-13}
				& PSNR$\uparrow$ & SSIM$\uparrow$ & LPIPS$\downarrow$ & PI$\downarrow$ &NIQE$\downarrow$ &MUSIQ $\uparrow$  & PSNR$\uparrow$ & SSIM$\uparrow$ & LPIPS$\downarrow$  & PI $\downarrow$  &NIQE$\downarrow$ & MUSIQ $\uparrow$\\
				\hline
				input  &10.87 &0.73 &0.34 &3.98 &4.60 &71.21  &12.75 &0.56 &0.25 & 3.58 &4.03 & 67.54\\
				SCI-easy$^\star$ \cite{SCI2022}  &7.96 &0.67 &0.52 &4.55 &3.75 &66.16 &20.79&\textcolor{blue}{0.82} &\textcolor{red}{0.09} &3.28 &3.88 &69.93 \\
				SCI-medium$^\star$ \cite{SCI2022}  &6.73 &0.62 &0.66 &8.42 &11.13 &60.46 &14.06 &0.75 &0.17 &3.61 &4.22 &69.42 \\
				SCI-difficult$^\star$ \cite{SCI2022}  &8.10 &0.70 &0.47 &\textcolor{blue}{3.73} &4.90 &68.78 &18.35 &0.77 &0.14 &3.33 &4.16 &71.30 \\
				URetinex-Net \cite{URetinexNet}  &8.37 &0.68 &0.50 &4.64 &5.87 &66.67 &14.33 &0.76 &0.17 &4.16 &4.08 &\textcolor{brown}{71.47} \\			
				RUAS-5K$^\star$ \cite{RUAS2021}  &6.70 &0.55 &0.67 &10.04& - &60.04  &18.35 &0.77 &0.35  & 4.51& 5.51 &67.73 \\
				RUAS-LOL$^\star$ \cite{RUAS2021} &6.37 &0.54 &0.71 &12.75& - &57.92 &7.70 &0.54 &0.48  & 4.66 & 6.41 &67.43\\
				Zhao et al. \cite{INN} &12.98 &0.75 &0.34 &3.91&4.55  &71.19  &19.11 &0.74 &\textcolor{brown}{0.12} &3.16 &\textcolor{red}{3.63}   &70.24\\
				EnlightenGAN$^\dagger$ \cite{Jiang2019} &9.04 &0.68 &0.40 &4.10&4.57  &70.37 &14.60 &0.75 &0.15 & 3.28 &3.74  &71.31\\
				Zero-DCE$^\star$ \cite{Guo2020CVPR} &9.07 &0.68 &0.40 &3.93 &4.59  &70.13  &18.10 &\textcolor{brown}{0.78} &\textcolor{brown}{0.12}  &\textcolor{brown}{3.14} &3.85   &\textcolor{blue}{72.19}\\
				DRBN \cite{Yang2020CVPR}&8.79 &0.65 &0.48 &4.88& 5.40  &66.90 &16.53  &0.77 &0.25 & 3.74 & 3.97  &70.94\\
				Retinex-Net \cite{Chen2018}&9.69 &0.70 &0.38 &\textcolor{brown}{3.76}& 4.38  &71.02  &12.47 &0.60 &0.25 & \textcolor{blue}{3.13} &  4.27 &\textcolor{red}{73.74}\\
				HDRNet \cite{gharbi2017deep}&  9.76& 0.72 & 0.44 &4.15 &5.00    &57.35  &\textcolor{blue}{21.22}  &\textcolor{brown}{0.79}  &\textcolor{blue}{0.11}  &3.41   & 3.93   &56.49 \\
                    Adaptive LUT \cite{zeng2020learning}&15.38  &\textcolor{blue}{0.80}  & 0.31 & 3.90& 4.42   &62.26  &20.00  & 0.75 &\textcolor{blue}{0.11}  &3.57   & 4.17   &56.85 \\
				\hline
				\textit{Afifi et al.}  \cite{Mahoud2021}&\textcolor{blue}{17.82} &\textcolor{blue}{0.80} &\textcolor{brown}{0.27} &\textcolor{red}{3.66}&\textcolor{red}{4.00} &  \textcolor{blue}{72.30} &\textcolor{brown}{21.21} &0.78 &\textcolor{brown}{0.12} & 3.52 &3.85   &70.69 \\
				\textit{ExCNet}$^\star$ \cite{ExCNet} &11.96 &0.74 &0.32 &3.85 &4.55   &71.42 &17.69 &0.77 &\textcolor{blue}{0.11} & \textcolor{red}{3.07} & \textcolor{brown}{3.67} &\textcolor{blue}{72.19} \\
				\textit{Exposure} \cite{Exposure} &16.35 &\textcolor{brown}{0.78} &0.32 &3.90 & 4.71  &72.15 &18.50 &0.72 &0.17  &  3.54 & 4.39  &69.64\\
				\hline
				Ours-auto$^\star$   &\textcolor{brown}{16.55}& \textcolor{red}{0.81} &\textcolor{red}{0.25}&3.92 &\textcolor{brown}{4.23}   &\textcolor{brown}{72.20} &19.76 &\textcolor{blue}{0.82} &\textcolor{brown}{0.12}  & 3.21 & \textcolor{blue}{3.66} &70.99 \\
				Ours-manu$^\star$  &\textcolor{red}{17.94} &\textcolor{red}{0.81} &\textcolor{blue}{0.26}  &\textcolor{brown}{3.76} & \textcolor{blue}{4.16}   &\textcolor{red}{72.53} &\textcolor{red}{22.49} &\textcolor{red}{0.84} &\textcolor{blue}{0.11}  &\textcolor{blue}{3.13} &\textcolor{red}{3.63}  &70.81\\
				\hline
		\end{tabular}}
	\end{center}
	\label{table:quantitative}
		\vspace{-1.2em}
\end{table*}

\noindent
\textbf{Visual Comparisons.}
We first show the results of different methods on overexposed images in Figures \ref{fig:overexposed_unpair1},  \ref{fig:overexposed_unpair2}, and  \ref{fig:overexposed_pair}. 
%
For the overexposed portraits as shown in   Figures \ref{fig:overexposed_unpair1} and \ref{fig:overexposed_unpair2},  Adaptive LUT \cite{zeng2020learning}, Zhao et al. \cite{INN}, Afifi et al. \cite{Mahoud2021}, and ExCNet \cite{ExCNet} produce relatively good results. 
%
Among them, Afifi et al. \cite{Mahoud2021} can correct the overexposure, but it introduces color deviation and artifacts in the results.
In contrast, our method corrects the overexposed regions with visually pleasing visual quality. 
In Figures \ref{fig:overexposed_pair}, Afifi et al. \cite{Mahoud2021} introduce artifacts while  the result of Zhao et al. \cite{INN} is still overexposed. 
%
In addition, Exposure \cite{Exposure} and Adaptive LUT \cite{zeng2020learning} change the color of the background when it is used to correct the overexposure. 
%
Compared with these methods, our result is closer to the reference image and looks more visually pleasing. 

\begin{table*}[!t]
	\caption{\textbf{Comparisons of FLOPs, trainable parameters, and runtime.} We denote methods designed for exposure correction in \textit{italic}. The runtime is computed on an NVIDIA 1080Ti GPU or Intel(R) Xeon(R) CPU@2.60GHz. The FLOPs are computed for processing an image of size 900$\times$1200$\times$3.  The best result is in \textcolor{red}{red} whereas the second and third are in \textcolor{blue}{blue} and  \textcolor{brown}{brown} under each case, respectively.  We discard ExCNet and Exposure for comparisons because their runtime and FLOPs change according to the input image content.}
	\begin{center}
		\resizebox{16cm}{!}{
			\begin{tabular}{c|c|c|c||c|c|c||c||c}
				\hline
				\multirow{2}{*}{Methods} & \multicolumn{3}{c||}{runtime on GPU$\downarrow$} & \multicolumn{3}{c||}{runtime on CPU$\downarrow$} & \multirow{2}{*}{FLOPs$\downarrow$} & \multirow{2}{*}{parameters$\downarrow$}\\
				\cline{2-7}
				\cline{2-7}
				& $256^2$ & $512^2$ & $1024^2$ & $2048^2$ & $3072^2$ & $4096^2$ &  \\
				\hline			
				SCI\cite{SCI2022}  &4.8ms &7.2ms &\textcolor{brown}{7.6ms} &\textcolor{brown}{0.3s} &\textcolor{blue}{0.5s} &\textcolor{blue}{0.8s} &\textcolor{brown}{350M} & \textcolor{red}{258} \\
				URetinex-Net \cite{URetinexNet}  &300.1ms &361.4ms &476.2ms &32.4s  &1.2min &2.1min &941G  &340K  \\	
				RUAS \cite{RUAS2021}   & 5.3ms &7.3ms &33.6ms &3.9s & 8.6s & 13.3s & 4G &  \textcolor{blue}{3K} \\
				Zhao et al. \cite{INN}  &186.2ms & 727.2ms &2.9s & 14.3min &31.3min & 56.0min &12T &12M   \\
				EnlightenGAN \cite{Jiang2019} &35.0ms  &48.7ms &67.3ms &26.6s &38.9s & 72.4s & 273G & 9M  \\
				Zero-DCE \cite{Guo2020CVPR} & 5.5ms &6.7ms & 28.6ms &19.6s &44.1s & 78.7s& 85G &  \textcolor{brown}{80K}  \\
				DRBN \cite{Yang2020CVPR} & 37.1ms & 153.0ms & 671.7ms  & 3.4min & 6.8min & 11.9min &196G & 580K   \\
				Retinex-Net \cite{Chen2018} & 57.8ms & 86.3ms& 227.7ms & 1.9s &4.0s & 9.3s &588G & 550K   \\
    		    HDRNet \cite{gharbi2017deep} &  \textcolor{blue}{1.5ms} & \textcolor{blue}{3.4ms} & 10.8ms  & 1.6s & 3.7s & 6.1s &\textcolor{blue}{103M} & 483K   \\
        		Adaptive LUT \cite{zeng2020learning} &  \textcolor{red}{0.8ms} & \textcolor{red}{1.1ms} & \textcolor{red}{1.8ms}  & \textcolor{red}{3.8ms} & \textcolor{red}{4.5ms} & \textcolor{red}{4.9ms} &\textcolor{red}{76M} & 594K  \\
				\hline
				\textit{Afifi et al.} \cite{Mahoud2021} &12.1ms &13.2ms &27.9ms &4.3s & 9.1s& 16.2s & 76G &  7M   \\
				\hline
				Ours  & \textcolor{brown}{4.7ms} & \textcolor{brown}{5.2ms} & \textcolor{blue}{5.4ms} & \textcolor{blue}{0.2s} &\textcolor{brown}{0.6s} & \textcolor{brown}{1.0s} & 6G & \textcolor{blue}{3K}   \\
				\hline
			\end{tabular}
		}
	\end{center}
	\label{table:Complex}
		\vspace{-1em}
\end{table*}

\begin{figure*}[!t]
	\centering
	\includegraphics[width=0.95\linewidth]{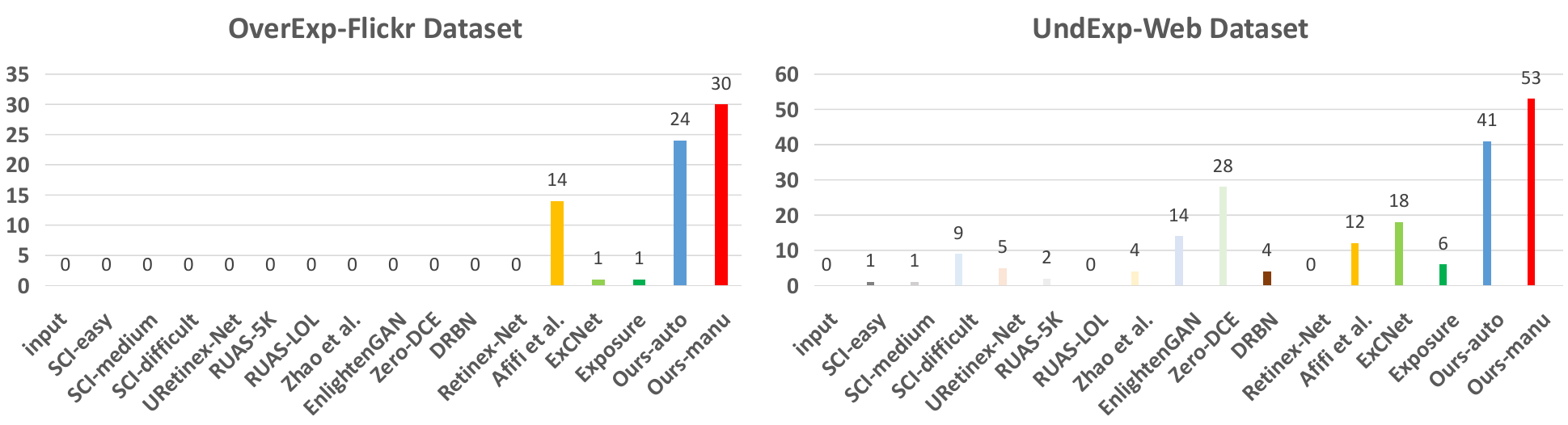}
	\caption{\textbf{User study on OverExp-Flickr and UndExp-Web datasets.}}
	\label{fig:statistics}
\end{figure*}

\begin{figure*}[!t]
	\centering
	\includegraphics[width=0.95\linewidth]{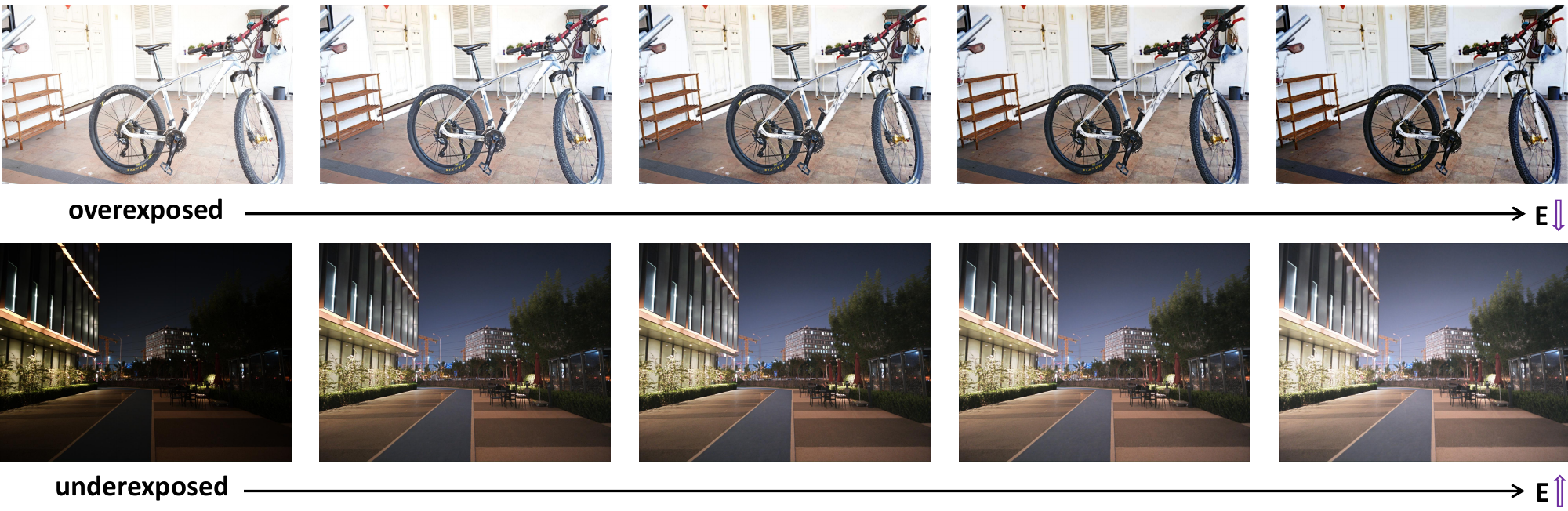}
	\caption{\textbf{Results of exposure control of our method.} For the overexposed input, the multiple results are generated by setting different uniform
		exposure	values (i.e., 0.4, 0.3, 0.2, 0.1) to the condition exposure map. For the underexposed input, the multiple results are generated by setting different uniform
		exposure	values (i.e., 0.5, 0.55, 0.6, 0.65) to the condition exposure map.  Our method is flexible in its controllable exposure adjustment.}
	\label{fig:diverse1}
	\vspace{-1em}
\end{figure*}

\begin{figure}[!t]
	\centering
	\includegraphics[width=1\linewidth]{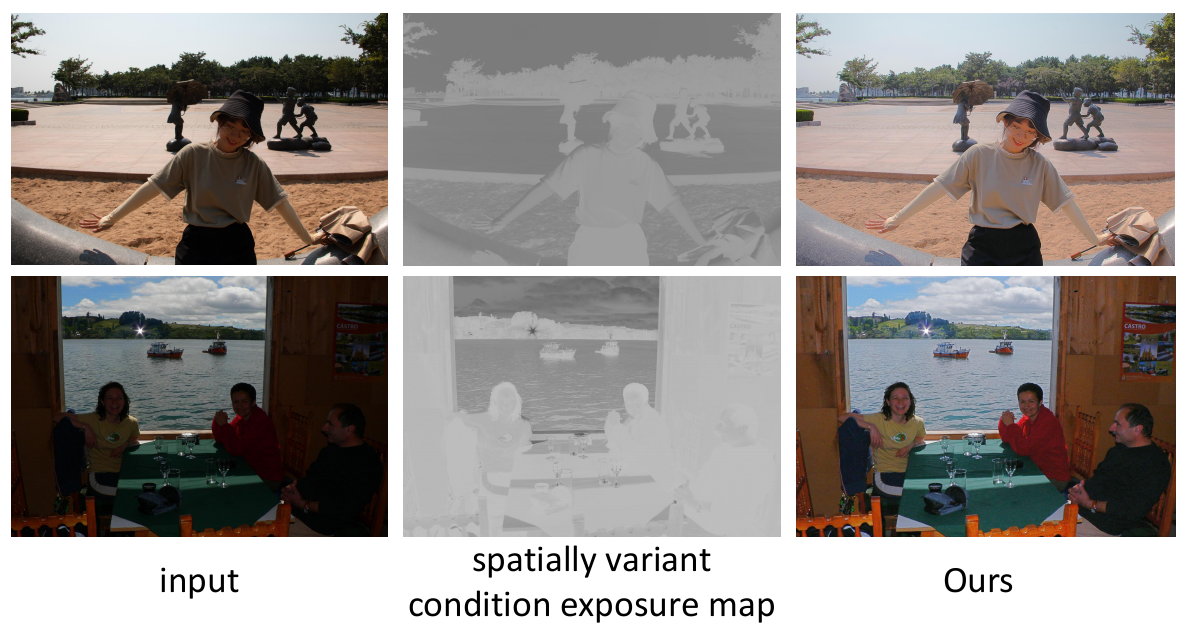}
		\vspace{-1em}
	\caption{\textbf{Results of spatially variant condition exposure map of our method.} The adjusted results are generated by setting the spatially variant exposure values to the different regions of condition exposure maps.}
	\label{fig:diverse2}
	\vspace{-1.3em}
\end{figure}

We then compare different methods on a challenging underexposed image that has saturated regions (e.g., the light sources) in Figure \ref{fig:underexposed_unpair}.
RUAS \cite{RUAS2021} introduces over-exposure and color deviation. 
%
HDRNet \cite{gharbi2017deep}, Adaptive LUT \cite{zeng2020learning}, Zhao et al. \cite{INN}, Afifi et al. \cite{Mahoud2021}, and Exposure \cite{Exposure} suffer from underexposure in some regions. 
%
EnlightenGAN \cite{Jiang2019} cannot cope with the saturation regions well.
Retinex-Net \cite{Chen2018} introduces artifacts. 
In contrast, our method effectively handles the underexposed and overexposed regions well and does not introduce artifacts.
Although we inherit the zero-reference learning and curve-based framework of Zero-DCE, our method differs from the original Zero-DCE because of the large teacher network and new self-supervised spatial exposure control loss.
Thus, our method achieves a better result than Zero-DCE.
Furthermore, we present the comparison results on another challenging underexposed image in Figure \ref{fig:underexposed_unpair2}. 
%
As presented, the compared methods tend to  produce color deviations and color over-saturation such as RUAS \cite{RUAS2021}, EnlightenGAN \cite{Jiang2019}, DRBN \cite{Yang2020CVPR},  Retinex-Net \cite{Chen2018}, and ExCNet \cite{ExCNet} or cannot enhance the local underexposed regions well, such as HDRNet \cite{gharbi2017deep}, Adaptive LUT \cite{zeng2020learning}, Zhao et al. \cite{INN}, Afifi et al. \cite{Mahoud2021}, and Exposure \cite{Exposure}.
%
Besides, some methods produce artifacts in their results, such as DRBN \cite{Yang2020CVPR} and Retinex-Net \cite{Chen2018}. Some methods produce over-enhancement such as URetinex-Net \cite{URetinexNet}, and RUAS \cite{RUAS2021}.
In contrast, our method achieves the most visually pleasing result with effective underexposure enhancement and does not introduce artifacts and artificial colors.

\noindent
\textbf{Quantitative Comparisons.}
For the testing datasets with reference images, we employ full-reference image quality assessment metrics PSNR, SSIM~\cite{SSIM}, and LPIPS \cite{LPIPS} to quantify the performance of different methods.
Otherwise, we employ the non-reference perceptual index (PI) \cite{PI}, naturalness image quality evaluator (NIQE) \cite{NIQE}, and multi-scale image quality Transformer (MUSIQ) (trained on the PaQ2PiQ dataset \cite{PaQ2PiQ}) to evaluate the quality of different results. 
The quantitative results are presented in Table \ref{table:quantitative}.

As summarized in Table \ref{table:quantitative}, `Ours-auto' already achieves compelling results that are comparable to the results of Afifi et al. \cite{Mahoud2021}, which is the state-of-the-art supervised exposure correction method.  `Ours-manu' outperforms all compared methods in terms of PSNR and SSIM for overexposure correction and underexposure correction, showing the flexibility and effectiveness of our approach.
Moreover, our method achieves comparable and even better non-reference image quality scores than the best performer under each case. 
The compared supervised methods in our experiments use publicly available pretrained models. Importantly, their training data are not drawn from the same distribution as our test datasets. Therefore, it is not surprising that their performance is suboptimal in our evaluation setting. For example, both HDRNet \cite{gharbi2017deep} and AdaptiveLUT \cite{zeng2020learning} were originally trained on the MIT-Adobe FiveK dataset. This dataset is significantly different from our test sets in terms of capture conditions and data distribution. As a result, although these models perform well on scenarios similar to their training data, their performance can degrade in unseen or domain-shifted settings. In contrast, zero-shot methods like our method, which rely less on paired supervision and more on intrinsic image statistics or self-consistency constraints, may demonstrate better adaptability in unseen or domain-shifted settings.
Please note that none of the methods, including ours, had access to the testing data during model training. Moreover, our method does not require any paired or unpaired training data and uses a single model to process all cases.
The quantitative comparisons suggest the good performance of our approach for coping with underexposed and overexposed images with a single model. 
This achievement is non-trivial in terms of a zero-reference learning-based method.

\noindent
\textbf{Computational Complexity Comparisons.}
In Table \ref{table:Complex}, we compare the computational complexity of different methods. Our method achieves the second and third fastest inference speed on GPU and CPU which is just a little slower than Adaptive LUT \cite{zeng2020learning}, SCI \cite{SCI2022}, and HDRNet \cite{gharbi2017deep}. In particular, our method has the advantage of processing high-resolution images. Moreover, the linear function of our method only involves Multiply-Add operation. 
SCI \cite{SCI2022}, RUAS \cite{RUAS2021}, Adaptive LUT \cite{zeng2020learning}, and HDRNet \cite{gharbi2017deep} are also efficient models, but they cannot cope with overexposed images well, and fail to enhance underexposed images in some cases, as shown in the visual and quantitative results. Note that Adaptive LUT \cite{zeng2020learning} uses C Programming Language to accelerate the inference speed.
Our method not only achieves good exposure adjustment but also has fast inference speed and a lightweight structure.  

\noindent
\textbf{User Study.}
For the testing datasets OverExp-Flickr and UndExp-Web without reference images, apart from the non-reference metrics, we also conduct a user study to measure the perceptual quality of the results of the underexposure enhancement and exposure correction methods.
Specifically, we invite 25 human subjects to independently vote for the results.
For each testing dataset, each human subject is shown 20 random groups of comparisons (each group includes the results processed by all compared methods) and votes for the preferred result in each group of comparison. 
The participants are required to choose the preferred result according to the following questions: 1) Is the result free of overexposure? 2) Is the result free of underexposure? 3) Is the result free of artifacts and unnatural texture? 4) Is the color of the result natural? 5) Is the result visually pleasing? 
We compute the number of votes for each method.
As shown in Figure \ref{fig:statistics}, the results of `Ours-auto' and `Ours-manu'  are preferred by participants for both underexposed and overexposed datasets, showing the good visual quality of our method.

\subsection{More Results}
\label{subsec: diverse}
In this part, we present our diverse results by controlling the condition exposure map. The diverse results of our method are achieved with a single model. The visual results are presented in Figures \ref{fig:diverse1} and \ref{fig:diverse2}.
As shown in Figure \ref{fig:diverse1}(top), by decreasing the exposure values, the overexposure of the overexposed images is adjusted, producing different exposure levels. Similarly, by increasing the exposure values, the brightness of the input underexposed image is gradually improved, as shown in Figure \ref{fig:diverse1}(bottom).

Besides using a uniform value for the condition exposure map, we can implement more elaborate exposure adjustment by setting spatially variant exposure values for different regions of the condition exposure map. 
By using the generation method of the spatially variant
condition exposure map introduced in Sec. \ref{exposuremap}, we assign the underexposed regions large exposure values and well-exposed/overexposed
regions small exposure values, as the spatially variant condition exposure maps suggested in Figure \ref{fig:diverse2}. 
For the final results, the underexposed regions are enhanced while the
well-exposed regions are well preserved.

\begin{figure}[!th]
	\centering
	\includegraphics[width=0.9\linewidth]{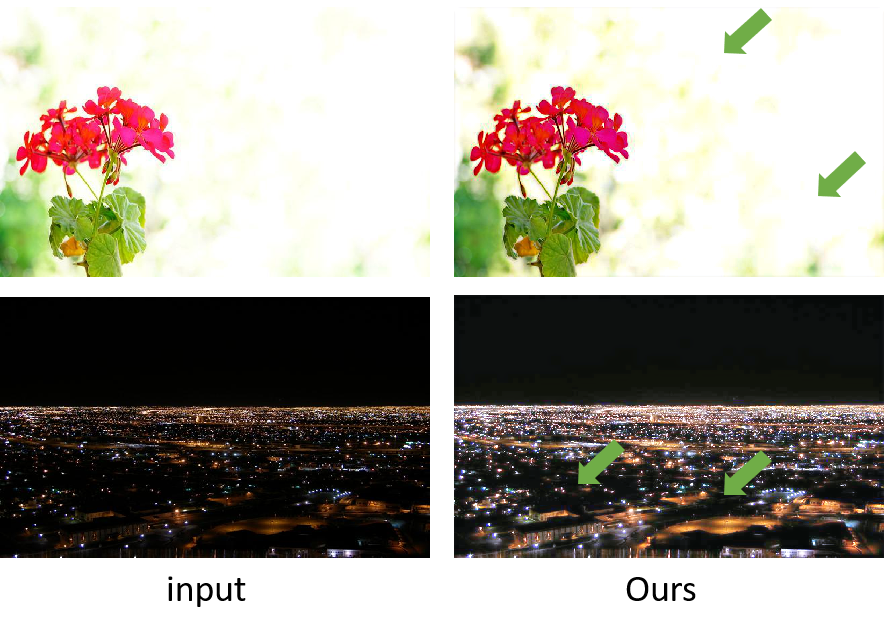}
	\caption{\textbf{Failure cases of our method.} Our method cannot cope with the extremely over-/under-exposed regions, in which the signal is clipped. The green arrows suggest the regions where our method cannot handle well.}
	\label{fig:failure}
	\vspace{-1.3em}
\end{figure}


\section{Conclusion}
We have presented a novel curve distillation method to accelerate existing curve-based estimation for exposure adjustment.  Apart from the new formulation, we have demonstrated that exposure adjustment is controllable with the proposed new self-supervised spatial exposure control loss.
Our method shows outstanding results on challenging underexposed and overexposed datasets despite the drastic speed up and lightweight structure.
Like existing methods, our method cannot cope with the extremely over-/under-exposed regions, in which the signal is clipped, as shown in Figure \ref{fig:failure}.
RAW data or image inpainting technologies may be a solution to these challenging cases. In addition, our method specifically focuses on exposure correction under monotonic constraints and does not explicitly model image noise. We assume that the input images are processed RGB images that have already undergone standard ISP procedures. Under this assumption, improper exposure is treated as the dominant degradation factor. Nevertheless, severely under-exposed regions may still exhibit low signal-to-noise ratio, and exposure enhancement without jointly considering noise may lead to noise amplification or reduced visual quality in these regions. Therefore, the current framework may have limitations when applied to extremely noisy low-light scenarios. Incorporating joint exposure correction and noise modeling could be an important direction for future work.

{
\bibliographystyle{IEEEtran}
\bibliography{egbib}
}


\begin{IEEEbiography}
[{\includegraphics[width=1in,height=1.25in,clip,keepaspectratio]{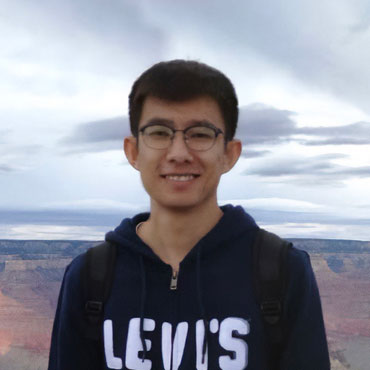}}]
{Chongyi Li} (Senior Member, IEEE) is a Professor with the School of Computer Science, Nankai University, China.  He was a  Research Assistant Professor with the S-Lab,  School of Computer Science and Engineering, Nanyang Technological University, Singapore from 2021 to 2023.  He was a Research Fellow with the City University of Hong Kong and Nanyang Technological University from 2018 to 2021. He received his Ph.D. degree from Tianjin University in 2018. He was also a joint training Ph.D. student at the Australian National University, Australia. His research interests include computer vision, generative AI, and computational imaging. 
\vspace{-4mm}
\end{IEEEbiography}

\begin{IEEEbiography}
[{\includegraphics[width=1in,height=1.25in,clip,keepaspectratio]{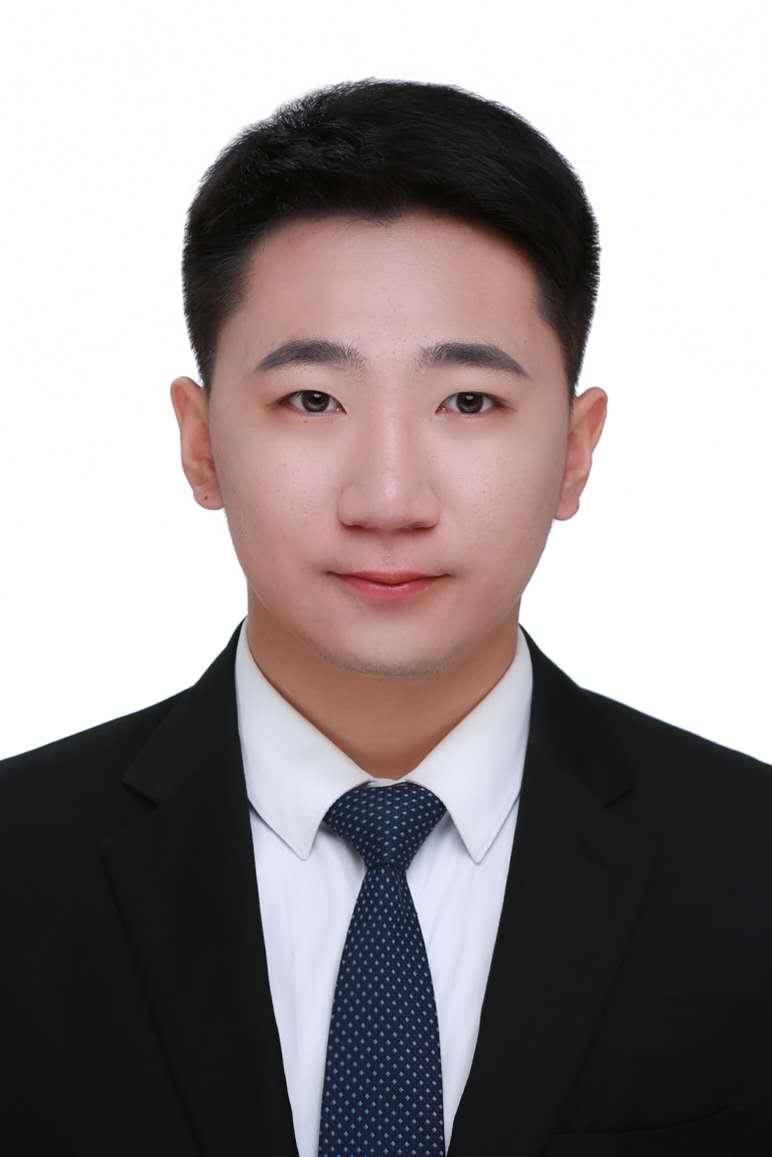}}]
{Chunle Guo}  (Member, IEEE) received a Ph.D. degree from Tianjin University, China, under the supervision of Prof. Jichang Guo. He was a Visiting Ph.D. Student with the School of Electronic Engineering and Computer Science, Queen Mary University of London (OMUL), U.K. He was a Research Associate with the Department of Computer Science, City University of Hong Kong (CityU). He was a Postdoctoral Researcher with Prof. Ming-Ming Cheng at Nankai University. He is currently an Associate Professor at Nankai University. His research interests include image processing, computer vision, and deep learning.
\vspace{-4mm}
\end{IEEEbiography}

\begin{IEEEbiography}[{\includegraphics[width=1in,height=1.25in,clip,keepaspectratio]{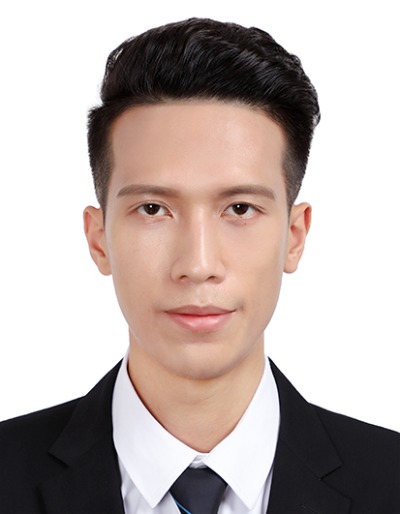}}]
{Ruicheng Feng} is a Senior Researcher with Hunyuan Team, Tencent. He received his Ph.D. degree from Nanyang Technological University in 2025. He was selected as an outstanding reviewer in NeurIPS 2023. He got the winner award in the NTIRE 2019 Real Image Super-Resolution challenges. He also co-organized the "MIPI workshop" series. His research interests lie in computational photography, image/video image generation, visual agentic AI, etc.
\vspace{-4mm}
\end{IEEEbiography}

\begin{IEEEbiography}[{\includegraphics[width=1in,height=1.25in,clip,keepaspectratio]{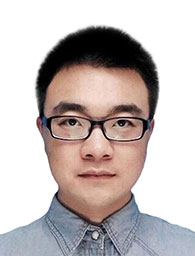}}]
{Shangchen Zhou} is currently a Research Assistant Professor at MMLab@NTU, Nanyang Technological University, Singapore. He received his Ph.D. (2024) in Computer Science from the same institution. He received the NTU CCDS Outstanding PhD Thesis Award in 2025. He won first place in three image restoration and enhancement challenges at NTIRE 2021. His works received notable recognition including the WAIC Youth Outstanding Paper Award Honorable Mention in 2023, the Snap Fellowship Honorable Mention in 2022, and the Best Paper Award at ICIMCS 2016. Additionally, he co-organized the “MIPI workshop” series in conjunction with ECCV 2022, CVPR 2023, CVPR 2024, and ICCV 2025. His research interests include image/video restoration and enhancement, generation and editing, etc.
\vspace{-4mm}
\end{IEEEbiography}

\begin{IEEEbiography}
[{\includegraphics[width=1in,height=1.25in,clip,keepaspectratio]{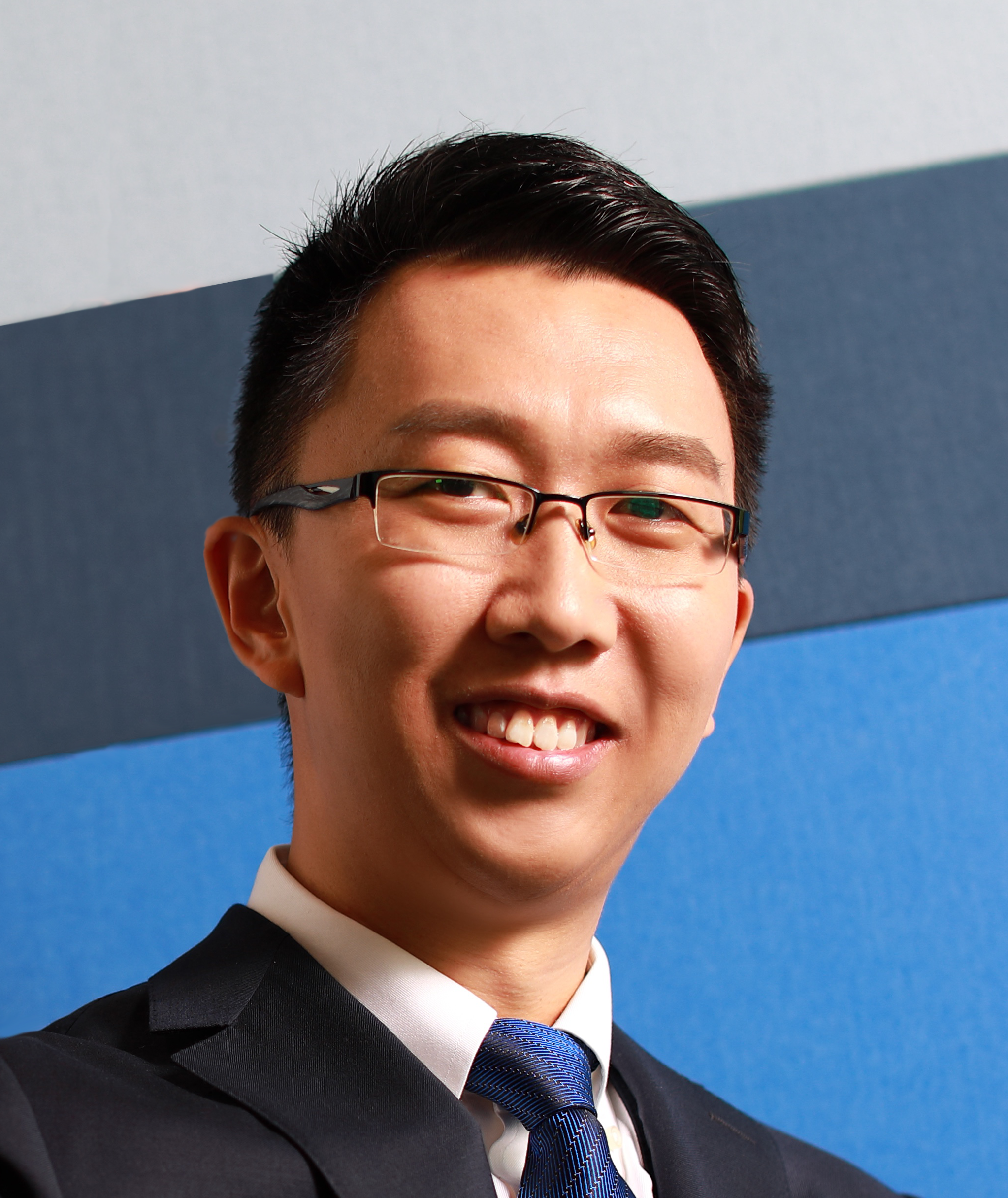}}]{Chen Change Loy}
(Senior Member, IEEE) is a President's Chair Professor at the College of Computing and Data Science, Nanyang Technological University (NTU), Singapore. He is the Director of MMLab@NTU and Co-Associate Director of S-Lab. He received his Ph.D. in Computer Science from Queen Mary University of London in 2010. Prior to joining NTU, he served as a Research Assistant Professor at the Multimedia Laboratory of The Chinese University of Hong Kong.
His research focuses on large multimodal models, generative AI, spatial intelligence and representation learning. Prof. Loy currently serves or has served as an Associate Editor for leading journals including the International Journal of Computer Vision (IJCV), IEEE Transactions on Pattern Analysis and Machine Intelligence (TPAMI), and Computer Vision and Image Understanding (CVIU). He has also served as an Area Chair or Senior Area Chair for major conferences such as CVPR, ICCV, ECCV, ICLR, and NeurIPS. He serves as Program Co-Chair of CVPR 2026 and General Co-Chair of ACCV 2028.
\end{IEEEbiography}

\vfill

\end{document}